\documentclass[a4paper,fleqn]{cas-dc}
\usepackage[numbers,sort&compress]{natbib}
\usepackage{amsmath,amsfonts,amssymb,amsthm}
\hypersetup{colorlinks=true,breaklinks=true,citecolor=blue,linkcolor=red}

\usepackage{graphicx}
\usepackage{booktabs}
\usepackage{multirow}
\usepackage{subcaption}
\usepackage{url}
\usepackage{xcolor}
\usepackage{balance}

\usepackage{algorithm}
\usepackage{algorithmic}
\usepackage{enumitem}

\usepackage{tikz}
\usepackage{pgfplots}
\pgfplotsset{compat=1.17}
\usetikzlibrary{arrows.meta,positioning,fit,backgrounds,calc}

\ExplSyntaxOn
\cs_gset:Npn \printorcid {}
\cs_gset:Npn \__first_head: {}
\cs_gset:Npn \__cas_head: {}
\cs_gset:Npn \__first_footerline: {}
\cs_gset:Npn \__first_foot:
{
  \parbox[t]{\textwidth}
  {
    \centering \thepage
  }
}
\cs_gset:Npn \__cas_foot:
{
  \parbox[t]{\textwidth}
  {
    \centering \thepage
  }
}
\ExplSyntaxOff
\definecolor{revbg}{RGB}{255,255,180}
\definecolor{revmbg}{RGB}{220,220,220}

\newcommand{\rev}[1]{#1}
\newcommand{\revm}[1]{#1}
\newcommand{\revt}[1]{#1}

\newcommand{\eg}{{\textit{e.g.}}}
\newcommand{\etal}{{\textit{et al.}}}

\newcommand{\KL}{\mathrm{KL}}
\DeclareMathOperator{\diag}{diag}

\newtheorem{theorem}{Theorem}
\newtheorem{lemma}{Lemma}
\newtheorem{proposition}{Proposition}

\newtheorem{definition}{Definition}
\newtheorem{remark}{Remark}

\newcommand{\teaserfigtopskip}{-0.5em}
\newcommand{\teasercaptionaboveskip}{1pt}
\newcommand{\teaserfigbottomskip}{-0.8em}

\begin{document}
\let\WriteBookmarks\relax
\def\floatpagepagefraction{1}
\def\textpagefraction{.001}

\shorttitle{Sinkhorn-CPD}
\shortauthors{Zhang et al.}

\title[mode=title]{Sinkhorn-CPD: Robust point cloud registration via unbalanced entropic optimal transport}

\author[1]{Jin Zhang}
\ead{jinzhang2022@buaa.edu.cn}

\author[2,4]{Mingyang Zhao}
\ead{zhaomingyang@amss.ac.cn}

\author[1]{Bing Liu}
\ead{liubingksy@buaa.edu.cn}

\author[3]{Xin Jiang}
\cormark[1]
\ead{jiangxin@buaa.edu.cn}

\affiliation[1]{organization={LMIB \& School of Mathematical Sciences, Beihang University}, city={Beijing}, country={China}}
\affiliation[2]{organization={State Key Laboratory of Mathematical Sciences, Academy of Mathematics and Systems Science, Chinese Academy of Sciences}, city={Beijing}, country={China}}
\affiliation[3]{organization={Beijing Key Laboratory of Artificial Intelligence Innovation and Application in the Machine Tool Industry, School of Artificial Intelligence, Beihang University}, city={Beijing}, country={China}}
\affiliation[4]{organization={University of Chinese Academy of Sciences}, city={Beijing}, country={China}}
\cortext[1]{Corresponding author}

\begin{abstract}
Coherent Point Drift (CPD) is widely used for rigid point cloud registration because of its soft correspondences and closed-form parameter updates.
However, CPD's target-side marginal constraint forces every observation, including outliers, to receive exactly unit probability mass.
This assumption degrades registration accuracy under heavy outliers and partial overlap.
Optimal transport (OT) methods can handle missing mass through unbalanced formulations, but require hand-tuned annealing schedules.
In this paper, we propose \emph{Sinkhorn-CPD}, which replaces CPD's target-side marginal constraint with dual Kullback--Leibler penalties, allowing the algorithm to discard outliers on both sides.
The resulting formulation is a fully unbalanced \emph{entropic optimal transport problem}, which can be efficiently solved by generalized Sinkhorn iterations. Moreover, Sinkhorn-CPD preserves the closed-form Procrustes and variance updates of CPD.
In our method, the variance $\sigma^2$ plays the role of the entropic regularization parameter, which induces an automatic annealing schedule from diffuse to sharp correspondences without manual temperature tuning.
Experiments on synthetic, cross-category, and scan-to-CAD benchmarks show that Sinkhorn-CPD achieves state-of-the-art accuracy, with strong robustness to outliers and partial overlap.
\end{abstract}


\maketitle

\noindent\textbf{Publication note.} The journal version of this work was published in \emph{Computer-Aided Design}: \url{https://doi.org/10.1016/j.cad.2026.104104}.

\noindent\textbf{Data availability.} Data and code are publicly available at \url{https://github.com/Theigrams/SinkhornCPD}.

\section{Introduction}
\label{sec:intro}

Rigid point cloud registration estimates a rotation $\mathbf{R}\in\mathrm{SO}(D)$ and translation $\mathbf{t}\in\mathbb{R}^{D}$ that aligns a source point set $\mathcal{Y}=\{\mathbf{y}_m\}_{m=1}^{M}\subset\mathbb{R}^{D}$ to a target point set $\mathcal{X}=\{\mathbf{x}_n\}_{n=1}^{N}$. 
Coherent Point Drift (CPD)~\cite{myronenko2010point} models the source
points as Gaussian mixture centroids and estimates the transformation by
maximizing the likelihood of the target points under the
Expectation-Maximization (EM) algorithm. 
The E-step computes soft correspondences as posterior responsibilities, and the M-step updates the transformation in closed form via Procrustes analysis.
All centroids share a single variance $\sigma^2$ that decreases as alignment improves, providing an intrinsic coarse-to-fine annealing mechanism.

\begin{figure}[pos=htp]
  \centering
  \includegraphics[width=\columnwidth]{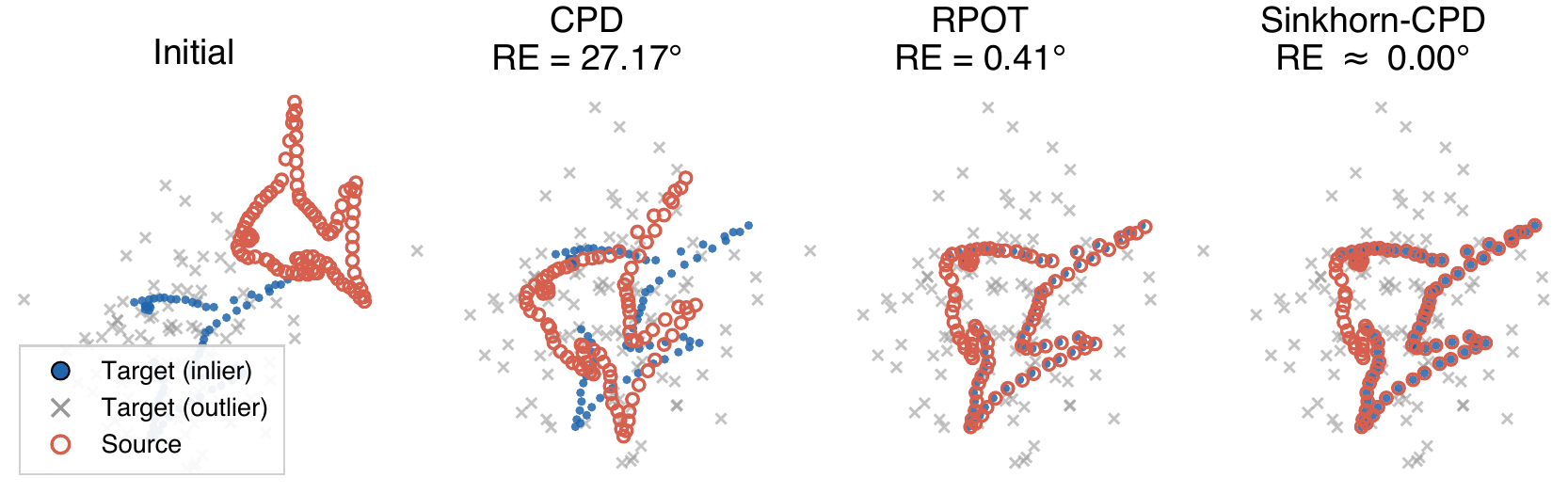}
  \caption{2D fish registration with $58.5\%$ target outliers.
  From left to right: initial configuration, CPD ($\mathrm{RE}{=}27.17^{\circ}$), RPOT ($\mathrm{RE}{=}0.41^{\circ}$), and Sinkhorn-CPD ($\mathrm{RE}{\approx}0.00^{\circ}$).
 CPD~\cite{myronenko2010point} suffers from severe misalignment caused by outliers. RPOT~\cite{qin2022partialot} mitigates this via partial transport yet converges slowly under fixed annealing schedules. Our method enables rapid, accurate alignment through dual KL-based outlier rejection and adaptive annealing.}
  \label{fig:teaser}
\end{figure}

However, CPD imposes a \emph{one-sided marginal constraint}:
the posterior responsibilities satisfy
\revt{$\sum_m P(m \mid \mathbf{x}_n) = 1$}, forcing every target point $\mathbf{x}_n$, including outliers and non-overlapping points, to distribute exactly unit probability mass across the source centroids.
Although a uniform outlier component absorbs some mass, under heavy outliers or low overlap the inclusion of non-corresponding points biases the Procrustes solution toward local minima. Fig.~\ref{fig:teaser} illustrates the effect: on a 2D fish shape with $58.5\%$ target outliers, CPD incurs a rotation error of $27.17^{\circ}$.

Optimal transport (OT) provides an alternative framework for computing soft correspondences.
Entropy-regularized OT and the Sinkhorn algorithm~\cite{cuturi2013sinkhorn} compute a transport plan between the two point sets, and unbalanced or partial OT variants~\cite{chizat2018uot,bonneel2019spot,qin2022partialot} relax the marginal constraints to discard unmatched mass.
However, existing OT-based registration methods typically adopt the squared Euclidean distance as the transport cost and rely on a prescribed annealing schedule for the entropy parameter~$\varepsilon$, which slows convergence and limits generalization across datasets.

In this work, we prove that CPD's E-step is equivalent to solving a one-sided entropic OT problem, and the variance $\sigma^2$ acts as the entropy temperature.
This equivalence reveals two things: (i) the one-sided marginal constraint is the structural cause of CPD's outlier sensitivity, and (ii) the EM variance update is a data-adaptive entropy annealing schedule.
Motivated by this analysis, we replace the one-sided hard marginal with dual-sided Kullback–Leibler (KL) penalties while retaining the CPD-induced Gaussian log-likelihood as the transport cost.

Building on this, we propose \emph{Sinkhorn-CPD}, which formulates rigid registration as an unbalanced entropic OT problem solved by generalized Sinkhorn iterations. It rejects outliers on both sides, retains the closed-form Procrustes and variance updates of CPD, and inherits automatic coarse-to-fine annealing through $\sigma^2$ without manual temperature tuning.

The main contributions are as follows:
\begin{itemize}
  \item We prove that CPD's E-step solves a one-sided entropic OT problem, identifying the variance $\sigma^2$ as the entropy temperature and the target marginal constraint as the structural cause of outlier sensitivity.
  \item  We cast rigid registration as an unbalanced entropic OT problem with dual-sided KL relaxations. The proposed Sinkhorn-CPD algorithm rejects outliers on both sides via generalized Sinkhorn iterations while maintaining closed-form updates.
  \item Experimental results show that Sinkhorn-CPD achieves state-of-the-art performance on synthetic data and real-world tasks, with strong robustness to outliers and partial overlap.
\end{itemize}

\section{Related Work}
\label{sec:related-work}

\subsection{ICP and Correspondence-Based Methods}

ICP~\cite{besl1992method} alternates between hard nearest-neighbor correspondences and least-squares rigid estimation.
Trimmed ICP~\cite{chetverikov2002trimmed} improves robustness to partial overlap by discarding point pairs with large residuals.
Sparse ICP~\cite{bouaziz2013sparseicp} replaces the least-squares objective with an $\ell_p$ ($p{\leq}1$) penalty to down-weight outliers.
Zhou~\etal~\cite{zhou2016fast} combined FPFH features~\cite{rusu2009fast} with a Geman--McClure objective for Fast Global Registration (FGR), avoiding iterative closest-point queries.
Yang~\etal~\cite{yang2021teaser} proposed TEASER++, which uses graph-based correspondence pruning and graduated non-convexity to tolerate extreme outlier ratios with certifiable guarantees.
These methods work well given a good initialization or distinctive local features, but their hard-assignment nature limits robustness under noise and low overlap.

\subsection{Probabilistic Registration}

Probabilistic methods replace hard correspondences with soft responsibilities.
Granger and Pennec~\cite{granger2002emicp} proposed multi-scale EM-ICP, which uses soft assignments within a coarse-to-fine hierarchy.
Jian and Vemuri~\cite{jian2010robust} represented both point sets as Gaussian mixtures and minimized their $L_{2}$ distance (GMMReg).
Myronenko and Song~\cite{myronenko2010point} introduced Coherent Point Drift (CPD), which models source points as isotropic Gaussian centroids with a uniform outlier component and updates the rigid transform and variance in closed form via EM.
Gao and Tedrake~\cite{gao2019filterreg} proposed FilterReg, which casts the E-step of probabilistic registration as a Gaussian filtering problem, achieving speed close to ICP while retaining probabilistic robustness.
Hirose~\cite{hirosem2020bcpd} reformulated CPD in a Bayesian framework with an acceleration scheme (BCPD).
Liu~\etal~\cite{liu2021lsgcpd} incorporated local surface geometry into CPD to improve feature discrimination (LSG-CPD).
Zhao~\etal~\cite{zhao2023accurate} proposed a consistent clustering method for cross-modality registration.


\subsection{Optimal Transport-Based Registration}

Optimal transport (OT) provides a principled framework for comparing and aligning distributions~\cite{villani2009ot,peyre2019computational}.
Cuturi~\cite{cuturi2013sinkhorn} showed that entropic regularization makes OT scalable and differentiable via the Sinkhorn algorithm~\cite{sinkhorn1967scaling}.
Bonneel and Coeurjolly~\cite{bonneel2019spot} proposed SPOT, which solves partial OT through sliced 1D projections.
Shen~\etal~\cite{shen2021accurate} proposed RobOT, which uses unbalanced OT as a robust matching layer to improve both optimization-based and learning-based registration.
Qin~\etal~\cite{qin2022partialot} formulated registration as partial OT with hard marginal constraints to control the matched fraction (RPOT).
Ma~\etal~\cite{ma2024dot} augmented the classical discrete OT model with orthogonal and diagonal matrix terms and proposed a relaxed variant for outlier handling.
On the theoretical side, Chizat~\etal~\cite{chizat2018uot} established the scaling algorithm framework for KL-penalized unbalanced OT, and Feydy~\etal~\cite{feydy2019interpolating} introduced Sinkhorn divergences that interpolate between OT and MMD, providing positive and convex geometric losses.

\subsection{Learning-Based Approaches}

Deep networks learn feature descriptors or direct pose regressors for registration.
Aoki~\etal~\cite{aoki2019pointnetlk} combined PointNet features with a Lucas--Kanade update (PointNetLK).
Wang and Solomon~\cite{wang2019deep,wang2019prnet} proposed DCP and PRNet, which learn correspondences and solve weighted Procrustes problems.
Yew and Lee~\cite{yew2020rpm} combined robust point matching with learned features (RPM-Net).
Qin~\etal~\cite{qin2023geotransformer} introduced GeoTransformer, which uses geometric attention for coarse-to-fine matching.


\section{Preliminaries}
\label{sec:preliminaries}

\subsection{Notation and Problem Formulation}
\label{subsec:notation}

We stack $\mathbf{x}_n$ and $\mathbf{y}_m$ into matrices $\mathbf{X}\in\mathbb{R}^{N\times D}$ and $\mathbf{Y}\in\mathbb{R}^{M\times D}$:
\begin{equation}
\mathbf{X} \triangleq
\begin{bmatrix}
\mathbf{x}_1^\top\\
\vdots\\
\mathbf{x}_N^\top
\end{bmatrix}\in\mathbb{R}^{N\times D},
\qquad
\mathbf{Y} \triangleq
\begin{bmatrix}
\mathbf{y}_1^\top\\
\vdots\\
\mathbf{y}_M^\top
\end{bmatrix}\in\mathbb{R}^{M\times D}.
\end{equation}
Rigid registration estimates $\mathbf{R}$ and $\mathbf{t}$ such that
\begin{equation}
  T_{\boldsymbol{\theta}}(\mathbf{y})
  = \mathbf{R}\,\mathbf{y} + \mathbf{t},
  \qquad
  \boldsymbol{\theta}
  = \{\mathbf{R},\mathbf{t}\},
  \label{eq:rigid-transform}
\end{equation}
aligns $T_{\boldsymbol{\theta}}(\mathcal{Y})$ to $\mathcal{X}$.
We use $\mathbf{1}_K\in\mathbb{R}^K$ to denote the all-ones vector.
The operators $\log$ and $\exp$ are applied elementwise to matrices, and
$\oslash$ denotes elementwise division.

\subsection{Coherent Point Drift Revisited}
\label{subsec:cpd-preliminaries}

\subsubsection{GMM Observation Model}
\label{subsubsec:gmm-registration}

CPD models target points as independent samples from a Gaussian mixture
whose centroids are the transformed source points.
For $m\in\{1,\dots,M\}$,
\begin{equation}
  p(\mathbf{x}_n \mid z_n{=}m,\,\boldsymbol{\Theta})
  = \mathcal{N}\!\bigl(\mathbf{x}_n \,\big|\,
      T_{\boldsymbol{\theta}}(\mathbf{y}_m),\;
      \sigma^2 \mathbf{I}_D\bigr),
  \label{eq:cpd-gaussian}
\end{equation}
where $z_n\in\{1,\dots,M{+}1\}$ is a latent assignment variable and
$\boldsymbol{\Theta}=(\boldsymbol{\theta},\sigma^2)$.
The $(M{+}1)$-th component is a uniform outlier density
$p(\mathbf{x}_n \mid z_n{=}M{+}1) = 1/|\Omega|$,
where $|\Omega|$ denotes the data volume.
The mixture weights are
\begin{equation}
  \pi_m =
  \begin{cases}
    (1-w)/M, & m = 1,\dots,M,\\
    w,       & m = M{+}1,
  \end{cases}
  \label{eq:cpd-weights}
\end{equation}
where $w\in[0,1)$ is a prior outlier ratio.
The complete marginal likelihood of observing $\mathbf{x}_n$ is
\begin{equation}
  p(\mathbf{x}_n \mid \boldsymbol{\Theta})
  = \sum_{m=1}^{M+1} \pi_m\;
    p(\mathbf{x}_n \mid z_n{=}m,\,\boldsymbol{\Theta}).
  \label{eq:cpd-likelihood}
\end{equation}


\subsubsection{Variational Free Energy}
\label{subsubsec:cpd-vfe}

CPD estimates $\boldsymbol{\Theta}$ by maximizing the data log-likelihood via EM, which is equivalent to minimizing the \emph{variational free energy} (also known as the negative ELBO)~\cite{neal1998view}:
\begin{equation}
  \mathcal{F}_{\mathrm{VFE}}(q,\boldsymbol{\Theta})
  = \int_{q(\mathbf{Z})} q(\mathbf{Z})\log \frac{q(\mathbf{Z})}{p(\mathbf{X},\mathbf{Z}\mid \boldsymbol{\Theta})}\mathrm{d}\mathbf{Z}.
  \label{eq:cpd-free-energy-def}
\end{equation}
In the E-step, $q(\mathbf{Z})$ is set to the posterior responsibilities:
\begin{equation}
  q(z_n=m) = p(z_n=m\mid \mathbf{x}_n,\boldsymbol{\Theta}^\mathrm{old}) \triangleq P_{mn},
\end{equation}
where $\mathbf{P}\in\mathbb{R}_+^{(M+1)\times N}$ denotes the responsibility
matrix and satisfies $\mathbf{P}^\top \mathbf{1}_{M+1} = \mathbf{1}_N$.

The joint log-likelihood decomposes as
$\log p(\mathbf{X},\mathbf{Z}\mid \boldsymbol{\Theta})
=\log p(\mathbf{X}\mid\mathbf{Z},\boldsymbol{\Theta})+\log p(\mathbf{Z})$,
where $p(\mathbf{Z})$ is determined by the fixed mixture weights $\boldsymbol{\pi}$ in~Eq.~\eqref{eq:cpd-weights}.
The negative log-likelihood term $-\log p(\mathbf{X}\mid \mathbf{Z},\boldsymbol{\Theta})$ can be viewed as a cost measuring the fit quality. Writing
$D_{mn}=\|\mathbf{x}_n - T_{\boldsymbol{\theta}}(\mathbf{y}_m)\|^2$, we define the cost matrix $\mathbf{\tilde{C}} \in \mathbb{R}^{(M+1)\times N}$ with entries:
\begin{equation}
\begin{aligned}
\tilde{C}_{mn}(\boldsymbol{\theta}, \sigma^2) &\triangleq -\log p(\mathbf{x}_n \mid z_n=m),\\
&=
\begin{cases}
\displaystyle
\frac{D_{mn}}{2\sigma^2}
+ \frac{D}{2}\log(2\pi\sigma^2),
& m \leq M,\\
\log |\Omega|, & m = M+1.
\end{cases}
\end{aligned}
\label{eq:cpd-cost}
\end{equation}

Substituting Eq.~\eqref{eq:cpd-cost} into Eq.~\eqref{eq:cpd-free-energy-def}, the energy becomes:
\begin{equation}
\mathcal{F}_{\mathrm{VFE}} = \langle \mathbf{P}, \mathbf{\tilde{C}} \rangle + \sum_{n=1}^N \sum_{m=1}^{M+1} P_{mn} \log P_{mn} - \langle \mathbf{P}\mathbf{1}_N, \log \boldsymbol{\pi} \rangle,
\label{eq:cpd-vfe-expanded}
\end{equation}
where $\langle\cdot,\cdot\rangle$ denotes the Frobenius inner product.
Minimizing~Eq.~\eqref{eq:cpd-vfe-expanded} with respect to $\mathbf{P}$ yields CPD's E-step, while minimizing it over $\boldsymbol{\Theta}$ gives the M-step.

\subsection{Optimal Transport Theory}
\label{subsec:ot}

We briefly recall discrete entropic OT in the form we will use for registration.

\subsubsection{Discrete Optimal Transport}
\label{subsubsec:discrete-ot}

We represent the source and target point clouds as discrete measures
\begin{equation}
  \mu = \sum_{m=1}^{M} b_m\,\delta_{\mathbf{y}_m},
  \qquad
  \nu = \sum_{n=1}^{N} a_n\,\delta_{\mathbf{x}_n},
  \label{eq:discrete-measures}
\end{equation}
where $\mathbf{b}\in\mathbb{R}_+^M$ and $\mathbf{a}\in\mathbb{R}_+^N$ are
mass vectors with
$\mathbf{b}^{\top}\mathbf{1}_M = \mathbf{a}^{\top}\mathbf{1}_N = 1$, and
$\delta_{\mathbf{x}}$ is the Dirac mass at $\mathbf{x}$.
By default we set $a_n = 1/N$ and $b_m = 1/M$.

The \emph{transport plan} is a matrix $\mathbf{\Gamma}\in\mathbb{R}_+^{M\times N}$ whose entry $\Gamma_{mn}$ gives the mass moved from $\mathbf{y}_m$ to $\mathbf{x}_n$\footnote{We adopt the convention $\Gamma_{mn}=\gamma(\mathbf{y}_m,\mathbf{x}_n)$ to maintain
consistency with CPD. Standard OT literature often
uses $\Gamma_{ij}=\gamma(\mathbf{x}_i,\mathbf{y}_j)$.}.
\revt{Throughout, the row index $m$ ranges over source points and the column index $n$ over target points, so the row marginal $\mathbf{\Gamma}\,\mathbf{1}_N\in\mathbb{R}^M$ pairs with the source mass $\mathbf{b}$, while the column marginal $\mathbf{\Gamma}^{\top}\mathbf{1}_M\in\mathbb{R}^N$ pairs with the target mass $\mathbf{a}$.}

The balanced feasible set is
\begin{equation}
  \Pi(\mathbf{b},\mathbf{a})
  = \bigl\{
      \mathbf{\Gamma}\geq\mathbf{0}
      \;:\;
      \mathbf{\Gamma}\,\mathbf{1}_N = \mathbf{b},\;\;
      \mathbf{\Gamma}^{\top}\mathbf{1}_M = \mathbf{a}
    \bigr\}.
  \label{eq:discrete-marginals}
\end{equation}
Given a cost matrix $\mathbf{C}\in\mathbb{R}^{M\times N}$ with
$C_{mn}=c(\mathbf{y}_m,\mathbf{x}_n)$, balanced OT solves:
\begin{equation}
  \min_{\mathbf{\Gamma}\in\Pi(\mathbf{b},\mathbf{a})}
  \;\langle \mathbf{C},\,\mathbf{\Gamma}\rangle.
  \label{eq:discrete-ot}
\end{equation}
In most OT-based registration papers~\cite{bonneel2019spot,shen2021accurate,qin2022partialot,ma2024dot}, the cost uses squared Euclidean distance:
$C_{mn}=\|\mathbf{x}_n - T_{\boldsymbol{\theta}}(\mathbf{y}_m)\|^2$.

\subsubsection{Entropy-Regularized OT}
\label{subsubsec:entropic-ot}
Exact OT has super-cubic complexity, which makes it expensive for large point clouds.
Entropic OT~\cite{cuturi2013sinkhorn} adds the negative \emph{Shannon entropy} as a regularizer:
\begin{equation}
  \mathcal{H}(\mathbf{\Gamma})
  \triangleq
  \sum_{m=1}^{M}\sum_{n=1}^{N}
  {\Gamma}_{mn}\bigl(\log {\Gamma}_{mn}-1\bigr).
\end{equation}
The resulting objective is strictly convex and has a unique minimizer~\cite{cuturi2013sinkhorn}:

\begin{definition}[Entropic OT]
\label{def:entropic-ot}
For marginals $\mathbf{a},\mathbf{b}$, cost $\mathbf{C}$, and
regularization strength $\varepsilon>0$, the entropic OT problem is
\begin{equation}
  \min_{\mathbf{\Gamma}\in\Pi(\mathbf{b},\mathbf{a})}
  \;\langle \mathbf{C},\,\mathbf{\Gamma}\rangle
  + \varepsilon\,\mathcal{H}(\mathbf{\Gamma}).
  \label{eq:entropic-ot}
\end{equation}
\end{definition}
The parameter $\varepsilon$ serves two purposes: (i) it makes the problem strongly convex
so the solution is unique, and (ii) it controls the sharpness of the transport
plan. Large values produce diffuse plans while small values recover plans close to the sparse linear programming (LP) solution.

\begin{proposition}[Sinkhorn scaling~\cite{sinkhorn1967scaling,cuturi2013sinkhorn}]
\label{prop:entropic-scaling}
Define the \emph{Gibbs kernel}
$\mathbf{K}\in\mathbb{R}_+^{M\times N}$ by
$K_{mn}=\exp(-C_{mn}/\varepsilon)$.
The unique minimizer of~\eqref{eq:entropic-ot} has the factored form
\begin{equation}
  \mathbf{\Gamma}^{\star}
  = \diag(\mathbf{u})\;\mathbf{K}\;\diag(\mathbf{v}),
  \label{eq:sinkhorn-scaling}
\end{equation}
where $\mathbf{u}\in\mathbb{R}_+^M$ and $\mathbf{v}\in\mathbb{R}_+^N$ are
the unique positive scaling vectors satisfying
\begin{equation}
  \mathbf{u} = \mathbf{b}\oslash(\mathbf{K}\,\mathbf{v}),
  \qquad
  \mathbf{v} = \mathbf{a}\oslash(\mathbf{K}^{\top}\mathbf{u}).
  \label{eq:sinkhorn-iteration}
\end{equation}
Alternating these two updates yields the Sinkhorn algorithm.
\end{proposition}


\section{The Sinkhorn-CPD Framework}
\label{sec:method}

In this section, we formulate rigid point cloud registration as an unbalanced entropic optimal transport problem. We first analyze the Expectation-Maximization (EM) steps of CPD from an optimal transport perspective to pinpoint the structural cause of its outlier sensitivity. We then introduce Sinkhorn-CPD, a dual-KL unbalanced formulation that enables symmetric outlier rejection while preserving the automatic variance annealing of CPD.

\subsection{Rethinking CPD: A One-Sided OT Perspective}
\label{subsec:cpd-as-ot}

Recall the CPD variational free energy $\mathcal{F}_{\mathrm{VFE}}$ under the $M$-component Gaussian mixture reduction (Section~\ref{subsubsec:cpd-vfe}). Let $\mathbf{D} \in \mathbb{R}_+^{M \times N}$ denote the squared Euclidean distance matrix with $D_{mn}=\|\mathbf{x}_n - T_{\boldsymbol{\theta}}(\mathbf{y}_m)\|^2$. We can rewrite the CPD E-step strictly in terms of $\mathbf{D}$.

\begin{theorem}[CPD E-step as one-sided entropic OT]
\label{thm:equivalence}
Assume uniform mixture weights $\pi_m=1/M$ and no outlier component ($w=0$). Up to an additive constant, minimizing the CPD E-step objective over the responsibility matrix $\mathbf{P} \in \mathbb{R}_+^{M \times N}$ subject to the column constraint $\mathbf{P}^{\top}\mathbf{1}_M = \mathbf{1}_N$ is equivalent to solving the one-sided entropic optimal transport problem:
\begin{equation}
  \min_{\substack{\mathbf{P}\geq\mathbf{0}\\ \mathbf{P}^{\top}\mathbf{1}_M = \mathbf{1}_N}}\;
  \langle \mathbf{D},\,\mathbf{P}\rangle
  + 2\sigma^2\,\mathcal{H}(\mathbf{P}),
  \label{eq:cpd-as-eot}
\end{equation}
where $\mathcal{H}(\mathbf{P})=\sum_{m,n}P_{mn}(\log P_{mn}-1)$.
\end{theorem}
The proof is given in Appendix~\ref{app:cpd-ot}.

\begin{remark}
\revt{When $w>0$, the outlier row can be eliminated via the column constraint, yielding $\bar{\mathbf{P}}^{\top}\mathbf{1}_M \leq \mathbf{1}_N$. The one-sided structure is unchanged: every target point still allocates its full unit mass, so the asymmetry persists for any $w\geq 0$.}
\end{remark}

This explains why CPD has difficulty handling target-side outliers. The hard constraint $\mathbf{P}^{\top}\mathbf{1}_M = \mathbf{1}_N$ (or $\bar{\mathbf{P}}^{\top}\mathbf{1}_M \leq \mathbf{1}_N$ when $w>0$) forces every target point $\mathbf{x}_n$, including severe outliers, to allocate its full unit probability mass. Source-side outliers safely dilute their mass because their row sums are free, but target-side outliers inject false correspondences that pull the rigid transformation into poor local minima.

Theorem~\ref{thm:equivalence} also clarifies CPD's annealing behavior. In Eq.~\eqref{eq:cpd-as-eot}, the regularization parameter is simply $\varepsilon = 2\sigma^2$, compared to standard entropic OT (Eq.~\eqref{eq:entropic-ot}). As EM iterations tighten the alignment and $\sigma^2$ shrinks, the entropy temperature drops automatically. The method moves from coarse global search to tight local matching without the manual temperature schedules that other OT-based methods require.

\subsection{Dual-KL Unbalanced Formulation}
\label{subsec:dual-kl}

To eliminate the target-side outlier trap while preserving the automatic annealing, we replace the hard target marginal with soft KL penalties on \emph{both} marginals.
\revt{This relaxation turns the balanced optimal transport of~\eqref{eq:entropic-ot} into an \emph{unbalanced} problem.}

\revt{Equivalently to using the raw squared distance with entropy weight $\varepsilon=2\sigma^2$ (Theorem~\ref{thm:equivalence}), we absorb the temperature into a normalized cost and fix the entropy coefficient to one.}
We define uniform prior marginals $\mathbf{a} = \frac{1}{N}\mathbf{1}_N$ and $\mathbf{b} = \frac{1}{M}\mathbf{1}_M$. The normalized cost matrix $\mathbf{C} \in \mathbb{R}^{M \times N}$ is derived from the Gaussian log-likelihood:
\begin{equation}
  C_{mn}(\mathbf{R},\mathbf{t},\sigma^2)
  = \frac{\|\mathbf{x}_n - T_{\boldsymbol{\theta}}(\mathbf{y}_m)\|^2}
         {2\sigma^2}
  + \frac{D}{2}\log(2\pi\sigma^2).
  \label{eq:cost-rigid}
\end{equation}
\revt{Since $\mathbf{C}$ scales as $1/\sigma^2$, shrinking $\sigma^2$ makes the cost term dominate over the fixed-scale entropy and KL terms, progressively sharpening the transport plan from diffuse global search to tight local matching.}

The Sinkhorn-CPD objective jointly optimizes the transport plan $\mathbf{\Gamma} \in \mathbb{R}_+^{M \times N}$ and the transformation parameters $\boldsymbol{\Theta}=(\mathbf{R},\mathbf{t},\sigma^2)$:
\begin{equation}
\begin{aligned}
  \min_{\mathbf{\Gamma} \geq \mathbf{0}, \boldsymbol{\Theta}} \mathcal{J}
  \;=\;&
  \langle \mathbf{C},\,\mathbf{\Gamma}\rangle
  + \mathcal{H}(\mathbf{\Gamma}) \\
  &+ \tau_x\,\KL\!\bigl(
      \mathbf{\Gamma}^{\top}\mathbf{1}_M
      \,\big\|\,\mathbf{a}
    \bigr)
  + \tau_y\,\KL\!\bigl(
      \mathbf{\Gamma}\,\mathbf{1}_N
      \,\big\|\,\mathbf{b}
    \bigr),
\end{aligned}
\label{eq:uot-objective}
\end{equation}
where $\tau_x,\tau_y > 0$ control the marginal relaxation strength, and $\KL(\mathbf{p}\|\mathbf{q}) = \sum_k(p_k\log\frac{p_k}{q_k} - p_k + q_k)$.
\revt{Here $\boldsymbol{\Gamma}$ is normalized so that $\sum_{m,n}\Gamma_{mn}\approx 1$, relating to CPD's responsibility matrix by $\boldsymbol{\Gamma}=\mathbf{P}/N$.}

By penalizing deviations from the uniform priors rather than strictly enforcing them, the objective allows outlier points on either side to shed their transported mass. If a target point lacks a geometric match, its column sum in $\mathbf{\Gamma}$ naturally approaches zero.

\subsection{Alternating Optimization}
\label{subsec:alternating}

We minimize $\mathcal{J}$ by alternating between a transport update (E-step) and a transformation update (M-step).

\subsubsection{E-Step: Generalized Sinkhorn Iterations}
\label{subsubsec:estep}

For fixed parameters $\boldsymbol{\Theta}$, solving Eq.~\eqref{eq:uot-objective} for $\mathbf{\Gamma}$ is a strictly convex unbalanced OT problem.

\begin{proposition}[Transport update~\cite{chizat2018uot}]
\label{prop:dual-kl-sinkhorn}
Define the Gibbs kernel $\mathbf{K} = \exp(-\mathbf{C}) \in \mathbb{R}_+^{M\times N}$ elementwise. Let $\alpha = \frac{\tau_y}{\tau_y + 1}$ and $\beta = \frac{\tau_x}{\tau_x + 1}$. The unique minimizer takes the scaling form $\mathbf{\Gamma}^{\star} = \diag(\mathbf{u})\,\mathbf{K}\,\diag(\mathbf{v})$, where $\mathbf{u}\in\mathbb{R}_+^M$ and $\mathbf{v}\in\mathbb{R}_+^N$ are the fixed points of the generalized Sinkhorn iterations:
\begin{equation}
\begin{aligned}
  \mathbf{u}^{(\ell+1)}
    &= \bigl(
         \mathbf{b}\oslash(\mathbf{K}\,\mathbf{v}^{(\ell)})
       \bigr)^{\!\alpha}, \\
  \mathbf{v}^{(\ell+1)}
    &= \bigl(
         \mathbf{a}\oslash(\mathbf{K}^{\top}\mathbf{u}^{(\ell+1)})
       \bigr)^{\!\beta}.
\end{aligned}
\label{eq:dual-kl-updates}
\end{equation}
\end{proposition}
The parameters $\alpha, \beta \in (0,1)$ dampen the projection steps. Unlike balanced Sinkhorn iterations, points in non-overlapping regions only weakly attract mass and naturally fade out. We initialize $\mathbf{u}^{(0)}=\mathbf{1}_M, \mathbf{v}^{(0)}=\mathbf{1}_N$ and run $L$ inner iterations.

\subsubsection{M-Step: Weighted Procrustes and Variance Update}
\label{subsubsec:mstep}

Given the fixed transport plan $\mathbf{\Gamma}$, we update $(\mathbf{R},\mathbf{t})$ and $\sigma^2$ in closed form.

\noindent\textbf{Rigid transform.}
Given $\mathbf{\Gamma}$ and $\sigma^2$, we update $(\mathbf{R},\mathbf{t})$.
The optimal rotation and translation minimize the $\mathbf{\Gamma}$-weighted least-squares objective:
\begin{equation}
  \min_{\mathbf{R}\in\mathrm{SO}(D),\,\mathbf{t}\in\mathbb{R}^D}
  \sum_{m,n}
    \Gamma_{mn}\,
    \|\mathbf{x}_n - (\mathbf{R}\,\mathbf{y}_m + \mathbf{t})\|^2.
  \label{eq:weighted-ls}
\end{equation}
Let $\gamma = \sum_{m,n}\Gamma_{mn}$ denote the total transported mass. We compute the weighted centroids:
\begin{equation}
  \bar{\mathbf{x}} = \frac{\mathbf{X}^{\top}\mathbf{\Gamma}^{\top}\mathbf{1}_M}{\gamma},
  \qquad
  \bar{\mathbf{y}} = \frac{\mathbf{Y}^{\top}\mathbf{\Gamma}\,\mathbf{1}_N}{\gamma}.
  \label{eq:weighted-centroids}
\end{equation}
The cross-covariance matrix can be written as
\begin{equation}
    \mathbf{S}
    =
    \mathbf{X}^{\top}\mathbf{\Gamma}^{\top}\mathbf{Y}
    - \gamma\,\bar{\mathbf{x}}\,\bar{\mathbf{y}}^{\top}
    \in \mathbb{R}^{D\times D}.
    \label{eq:cross-cov}
\end{equation}

\begin{lemma}[Weighted Procrustes solution]
\label{prop:procrustes}
Let $\mathbf{S}=\mathbf{U}\mathbf{\Sigma}\mathbf{V}^{\top}$ be the singular value decomposition of $\mathbf{S}$.
Then the minimizer of~\eqref{eq:weighted-ls} is
\begin{equation}
\begin{aligned}
    \mathbf{R}
    &= \mathbf{U}\,\operatorname{diag}(1,\ldots,1,\det(\mathbf{U}\mathbf{V}^{\top}))\,\mathbf{V}^{\top}, \\
    \mathbf{t}
    &= \bar{\mathbf{x}} - \mathbf{R}\,\bar{\mathbf{y}}.
\end{aligned}
\label{eq:procrustes-update}
\end{equation}
\end{lemma}

The proof follows the standard closed-form absolute orientation analysis~\cite{horn1987closedform}.
Because outliers on either side contribute near-zero mass to $\mathbf{\Gamma}$, they are automatically excluded from the Procrustes fit.

\noindent\textbf{Variance update.}
We update the variance $\sigma^2$ by setting the partial derivative of $\mathcal{J}$ with respect to $\sigma^2$ to zero (see Appendix~\ref{app:variance-derivation}), yielding:
\begin{equation}
  \sigma^2
  = \frac{1}{D\,\gamma}
    \sum_{m=1}^M \sum_{n=1}^N
      \Gamma_{mn}\,
      \|\mathbf{x}_n - (\mathbf{R}\,\mathbf{y}_m + \mathbf{t})\|^2.
  \label{eq:sinkhorn-cpd-sigma}
\end{equation}
The variance update is the core mechanism of robust annealing. The estimated variance accurately reflects the noise level of the \emph{inliers alone}. As the alignment improves, the inlier residuals decrease, which reduces $\sigma^2$ and automatically sharpens the optimal transport plan.

\subsection{Algorithm Summary and Complexity}
\label{subsec:algorithm}

Algorithm~\ref{alg:sinkhorn-cpd} outlines the complete Sinkhorn-CPD procedure. 

\begin{algorithm}[htbp]
\caption{Sinkhorn-CPD}
\label{alg:sinkhorn-cpd}
\begin{algorithmic}[1]
\REQUIRE Source $\mathbf{Y}\in\mathbb{R}^{M\times D}$, target $\mathbf{X}\in\mathbb{R}^{N\times D}$;
  KL weights $\tau_x,\tau_y > 0$; Sinkhorn iterations $L$; tolerance $\epsilon$.
\STATE Initialize $\mathbf{R}\leftarrow\mathbf{I}_D$, $\mathbf{t}\leftarrow \revt{\bar{\mathbf{x}} - \mathbf{R}\bar{\mathbf{y}}}$, $\sigma^2\leftarrow \revt{\sigma_0^2}$.
\STATE Set $\mathbf{a}\leftarrow\tfrac{1}{N}\mathbf{1}_N$, $\mathbf{b}\leftarrow\tfrac{1}{M}\mathbf{1}_M$.
\STATE Compute exponents $\alpha\leftarrow\tfrac{\tau_y}{\tau_y+1}$, $\beta\leftarrow\tfrac{\tau_x}{\tau_x+1}$.
\REPEAT
  \STATE \textit{// E-step: Transport update}
  \STATE Compute cost $\mathbf{C}$ via Eq.~\eqref{eq:cost-rigid}, and $\mathbf{K}\leftarrow\exp(-\mathbf{C})$.
  \STATE Initialize $\mathbf{u}\leftarrow\mathbf{1}_M$, $\mathbf{v}\leftarrow\mathbf{1}_N$.
  \FOR{$\ell = 1,\ldots,L$}
    \STATE $\mathbf{u}\leftarrow \bigl(\mathbf{b}\oslash(\mathbf{K}\mathbf{v})\bigr)^{\alpha}$
    \STATE $\mathbf{v}\leftarrow \bigl(\mathbf{a}\oslash(\mathbf{K}^{\top}\mathbf{u})\bigr)^{\beta}$
  \ENDFOR
  \STATE $\mathbf{\Gamma}\leftarrow \diag(\mathbf{u})\,\mathbf{K}\,\diag(\mathbf{v})$
  
  \STATE \textit{// M-step: Transformation update}
  \STATE Compute $\bar{\mathbf{x}}, \bar{\mathbf{y}}$ via Eq.~\eqref{eq:weighted-centroids}; form cross-covariance $\mathbf{S}$.
  \STATE Update $\mathbf{R}, \mathbf{t}$ via Eq.~\eqref{eq:procrustes-update}.
  \STATE Update $\sigma^2$ via Eq.~\eqref{eq:sinkhorn-cpd-sigma}.
\UNTIL{\revt{$|\mathcal{J}^{(k)} - \mathcal{J}^{(k-1)}| < \epsilon$}}
\ENSURE Rigid transformation $(\mathbf{R},\mathbf{t})$ and transport plan $\mathbf{\Gamma}$.
\end{algorithmic}
\end{algorithm}

\noindent\textbf{Complexity.}
The computational cost per outer iteration is dominated by the pairwise distance calculation for $\mathbf{C}$ ($\mathcal{O}(MND)$) and the $L$ Sinkhorn iterations, which require matrix--vector multiplications of $\mathcal{O}(LMN)$. The SVD in the Procrustes step scales as $\mathcal{O}(D^3)$, which is negligible \revt{for $D=3$}. \revt{Letting $T$ denote the number of outer iterations, the overall complexity is $\mathcal{O}(TMN(D+L))$; in practice $T\approx 15\text{--}30$ on ModelNet40, matching} the same practical scalability as CPD.


\section{Experiments}
\label{sec:experiments}

We evaluate Sinkhorn-CPD across four scenarios: a 2D mechanistic validation, a controlled synthetic benchmark (Stanford Bunny), a large-scale cross-category dataset (ModelNet40), and a real-world scan-to-CAD dataset. We also provide ablation studies to validate our core design choices.

\subsection{Experimental Setup}
\label{sec:exp-setup}

\noindent\textbf{Baselines.}
We compare against \revt{eight} classical methods spanning three families:
(i)~\emph{correspondence-based}: Sparse-ICP~\cite{bouaziz2013sparseicp}\footnote{\url{https://github.com/OpenGP/sparseicp}},
FGR~\cite{zhou2016fast}\footnote{\url{https://github.com/isl-org/Open3D}}, and
TEASER++~\cite{yang2021teaser}\footnote{\url{https://github.com/MIT-SPARK/TEASER-plusplus}};
(ii)~\emph{probabilistic}: CPD~\cite{myronenko2010point} and
FilterReg~\cite{gao2019filterreg} (via \texttt{probreg})\footnote{\url{https://github.com/neka-nat/probreg}}, and
LSG-CPD~\cite{liu2021lsgcpd}\footnote{\url{https://github.com/ChirikjianLab/LSG-CPD}};
(iii)~\emph{OT-based}: RPOT~\cite{qin2022partialot}\footnote{\url{https://github.com/Hongxing-CQU/RPOT}}
\revt{and RobOT~\cite{shen2021accurate}\footnote{\url{https://github.com/uncbiag/robot}}}.
On ModelNet40, we additionally include two learning-based methods:
RPMNet~\cite{yew2020rpm}\footnote{\url{https://github.com/yewzijian/RPMNet}} and
GeoTransformer~\cite{qin2023geotransformer}\footnote{\url{https://github.com/qinzheng93/GeoTransformer}}, using official pre-trained weights.

\noindent\textbf{Metrics.}
We adopt three standard metrics:
\emph{rotation error} $\mathrm{RE}=\arccos\bigl((\mathrm{tr}(\mathbf{R}_{\text{est}}^{\top}\mathbf{R}_{\text{gt}})-1)/2\bigr)$ in degrees,
\emph{translation error} $\mathrm{TE}=\|\mathbf{t}_{\text{est}}-\mathbf{t}_{\text{gt}}\|_{2}$, and
\emph{RMSE} (root-mean-square point deviation after alignment) defined on source points as
$\mathrm{RMSE}=\sqrt{\frac{1}{M}\sum_{m=1}^M \|(\mathbf{R}_{\text{est}}\mathbf{y}_m+\mathbf{t}_{\text{est}})-(\mathbf{R}_{\text{gt}}\mathbf{y}_m+\mathbf{t}_{\text{gt}})\|_2^2}$.
On ModelNet40 we further report \emph{registration recall} (RR): the percentage of pairs satisfying $(\mathrm{RE}<1^\circ)\wedge(\mathrm{TE}<0.1)$.

\noindent\textbf{Implementation.}
We implement our method in Python, using the PyTorch library~\cite{paszke2017automatic} for GPU acceleration. All experiments run on the same hardware/software environment (see Table~\ref{tab:exp-env}). For all iterative methods, we set the maximum iterations to \revt{50}. 
Unless otherwise stated, each configuration is evaluated over 20 independent trials with different random seeds, and we report the mean $\pm$ standard deviation. All methods share the same test samples to ensure a fair comparison.
We measure wall-clock runtime per pair, excluding data loading and preprocessing.
Sinkhorn-CPD is configured with $\tau_x=\tau_y=1$ on ModelNet40 and the real scan-to-CAD benchmark; on the Stanford Bunny sweeps we additionally report $\tau_y=0.1$ as a more aggressive robustness mode (Section~\ref{sec:e1}) to quantify the trade-off between outlier rejection and rotational capture range. We use $L=20$ Sinkhorn iterations per outer step and set the initial variance $\sigma_{0}^{2} = \tfrac{1}{MND}\sum_{m,n}\|\mathbf{x}_n - \mathbf{y}_m\|^2$.
All baselines use their official default parameters; a per-method summary is given in Appendix~\ref{app:baseline-params}.

\begin{table}[pos=htp]
  \centering
  \footnotesize
  \caption{Hardware and software environment used for all experiments.}
  \label{tab:exp-env}
  \begin{tabular}{ll}
    \toprule
    \textbf{Component} & \textbf{Specification} \\
    \midrule
    CPU & Intel Core i5-12400F (6C/12T) \\
    GPU & NVIDIA GeForce RTX 4060 Ti (16 GB) \\
    Memory & 32 GB DDR4 \\
    OS & Ubuntu 22.04 LTS \\
    \midrule
    Python & 3.9.18 \\
    Open3D & 0.17.0 \\
    probreg & 0.3.8 \\
    MATLAB & 2022b \\
    \bottomrule
  \end{tabular}
\end{table}

\subsection{Mechanistic Validation on 2D Data}
\label{sec:e0}

We use a 2D example to show the effect of target-side outliers. The source $\mathcal{Y}$ is a 91-point fish contour. The target $\mathcal{X}$ contains 71 inliers (a $-60^{\circ}$ rotation and translation $(-2,-2)$ applied to the source) plus 100 Gaussian outliers, creating a $58.5\%$ outlier ratio.

As shown in Fig.~\ref{fig:teaser}, CPD is severely affected by outliers, leading to alignment failure. RPOT~\cite{qin2022partialot} discards some outlier mass and achieves near-perfect alignment, but its fixed geometric decay schedule ($\varepsilon \leftarrow 0.9\,\varepsilon$) causes slow convergence: even after 100 iterations, it has not fully converged. Sinkhorn-CPD discards outlier mass on both sides through dual KL penalties and converges quickly due to its data-adaptive annealing.

\noindent\textbf{Vote (influence) visualization.}

To explain the underlying mechanism, we analyze the per-target vote weights that drive the rigid Procrustes update. 
\begin{definition}[Vote Weights]
  In the M-step, both CPD and Sinkhorn-CPD solve a weighted Procrustes problem.
  The \emph{vote} of a target point $\mathbf{x}_n$ is the sum of its column entries in the correspondence matrix: \revt{$v_n = \sum_{m} \mathbf{P}_{mn}/\sum_{m,n} \mathbf{P}_{mn}$} for CPD and \revt{$v_n = \sum_{m} \Gamma_{mn}/\sum_{m,n} \Gamma_{mn}$} for Sinkhorn-CPD.
\end{definition}

The vote weight $v_n$ quantifies how much $\mathbf{x}_n$ influences the Procrustes fit. The target centroid $\boldsymbol{\mu}_x=\sum_n v_n\mathbf{x}_n$ is the weighted average of the target points. If outliers receive significant vote weight, they bias the centroid and the resulting poor alignment.

\begin{figure*}[pos=t]
  \vspace{\teaserfigtopskip}
  \centering
  \begin{subfigure}[t]{0.22\textwidth}
    \centering
    \includegraphics[width=\linewidth]{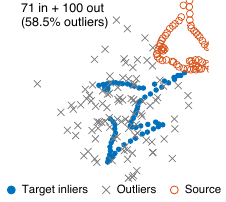}
    \caption{Initial setup.}
  \end{subfigure}\hfill
  \begin{subfigure}[t]{0.18\textwidth}
    \centering
    \includegraphics[width=\linewidth]{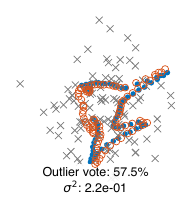}
    \caption{CPD result.}
  \end{subfigure}\hfill
  \begin{subfigure}[t]{0.18\textwidth}
    \centering
    \includegraphics[width=\linewidth]{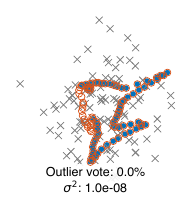}
    \caption{Sinkhorn-CPD result.}
  \end{subfigure}\hfill
  \begin{subfigure}[t]{0.35\textwidth}
    \centering
    \includegraphics[width=\linewidth]{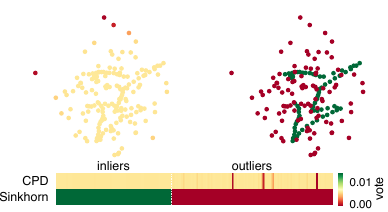}
    \caption{Vote weight distribution.}
  \end{subfigure}
  \captionsetup{font=footnotesize,aboveskip=\teasercaptionaboveskip}
  \caption{Toy 2D case with $58.5\%$ target outliers: Sinkhorn-CPD suppresses outlier votes and recovers near-zero rotation error.}
  \vspace{\teaserfigbottomskip}
  \label{fig:vote}
\end{figure*}

Fig.~\ref{fig:vote} compares the vote distributions of CPD and Sinkhorn-CPD. As shown in the heatmap strip~(d), CPD assigns $57.5\%$ of the total vote weight to outliers, which dominate the transformation. Sinkhorn-CPD suppresses outlier votes to $0.0\%$, concentrating the optimization strictly on inlier correspondences.

\subsection{Robustness on Synthetic Benchmark}
\label{sec:e1}

\begin{figure}[pos=htp]
  \centering
  \includegraphics[width=\columnwidth]{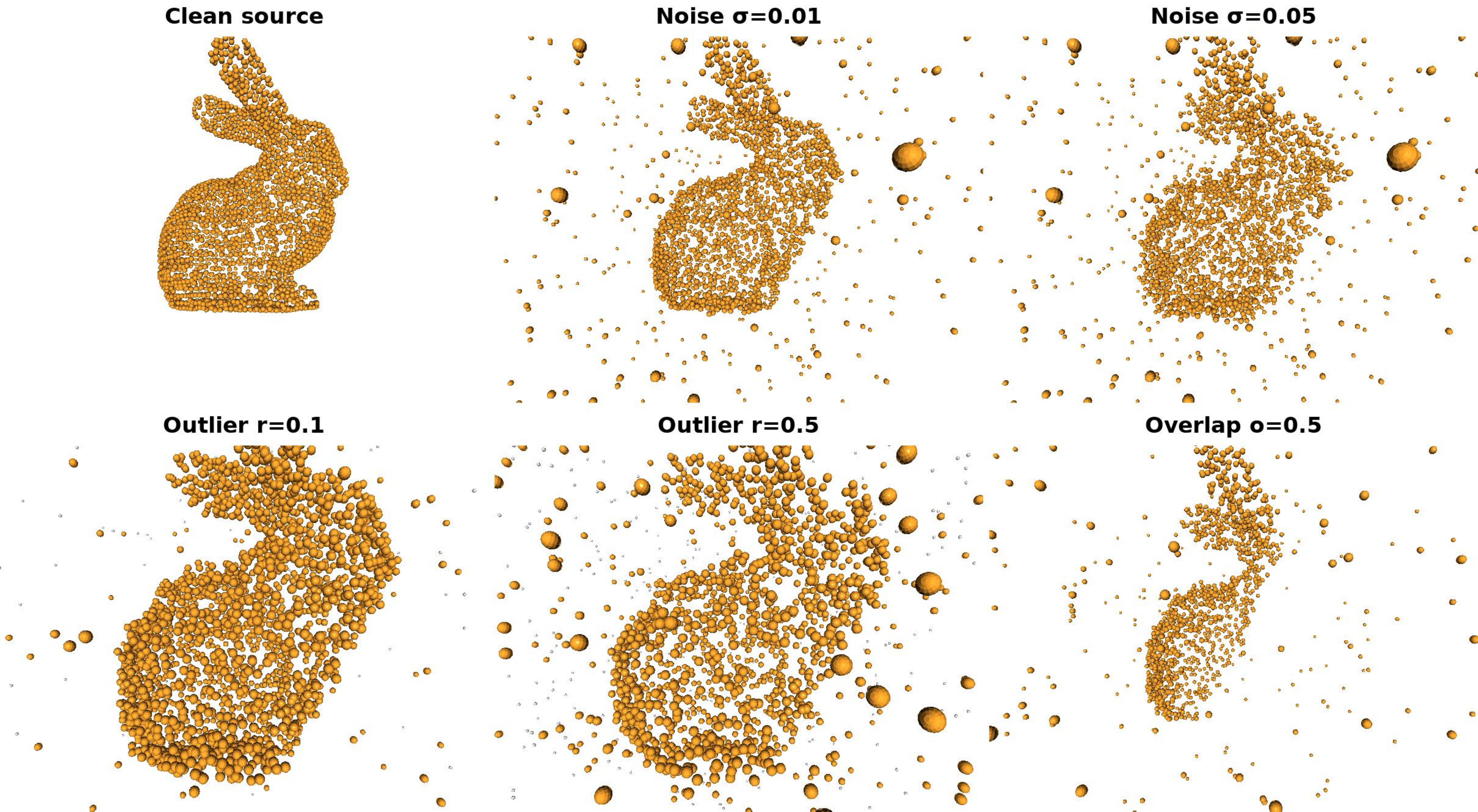}
  \caption{Perturbation examples on the Stanford Bunny.
  Top row: clean source, low noise ($\sigma{=}0.01$), high noise ($\sigma{=}0.05$).
  Bottom row: low outlier ratio ($r{=}0.1$), high outlier ratio ($r{=}0.5$), partial overlap ($o{=}0.5$).}
  \label{fig:perturbations}
\end{figure}

We evaluate robustness boundaries using the Stanford Bunny. We define four single-factor perturbation axes: \textbf{Noise}, \textbf{Outlier}, \textbf{Overlap}, and \textbf{Rotation}. For each axis, we vary one parameter while fixing the others at reference values (Table~\ref{tab:synthetic-protocol}).

\begin{table}[pos=htp]
    \centering
    \footnotesize
    \caption{Synthetic benchmark protocol. Each sub-experiment varies one factor while fixing the others.}
    \label{tab:synthetic-protocol}
    \begin{tabular}{lccc}
        \toprule
        Factor &  Parameter & Range & Fixed values \\
        \midrule
        Noise &
        $\sigma$ &
        $\{0.01,\dots,0.05\}$ &
        $\theta{=}30^{\circ}$, $r{=}0.2$, $o{=}0.9$ \\
        Outlier &
        $r$ &
        $\{0.1,\dots,0.7\}$ &
        $\theta{=}30^{\circ}$, $\sigma{=}0.02$, $o{=}0.9$ \\
        Overlap &
        $o$ &
        $\{0.4,\dots,0.9\}$ &
        $\theta{=}30^{\circ}$, $\sigma{=}0.02$, $r{=}0.2$ \\
        Rotation &
        $\theta$ &
        $\{10^{\circ},\dots,90^{\circ}\}$ &
        $\sigma{=}0.02$, $r{=}0.2$, $o{=}0.9$ \\
        \bottomrule
    \end{tabular}
\end{table}

\subsubsection{Data Generation}

All pairs are built from \texttt{bun\_zipper.ply} in the Stanford Bunny Dataset\footnote{\url{http://graphics.stanford.edu/data/3Dscanrep}}, centered and scaled to unit size, voxel-downsampled (voxel size $= 0.04$), and randomly subsampled to $N{=}3{,}000$ points.
For each trial we perturb the target in four ways (Fig.~\ref{fig:perturbations}):
\begin{enumerate}[nosep,leftmargin=1.5em]
  \item \revt{\textbf{Partial overlap.} Crop the target with a random half-space, keeping a fraction $o$ of points (the source stays intact).}
  \item \textbf{Noise.} Add Gaussian noise with standard deviation $\sigma_n$.
  \item \revt{\textbf{Outliers.} Replace a fraction $r$ of target points with uniform outliers from a ball of radius $2$.}
  \item \textbf{Rotation.} \revt{Rotate $\theta$ degrees around a random unit axis.}
\end{enumerate}

\subsubsection{Quantitative Results}

\noindent\textbf{Reference setting.}
Table~\ref{tab:e1-reference} reports RMSE and runtime under the reference configuration.

\begin{table}[pos=htp]
  \centering
  \footnotesize
  \caption{Reference configuration ($\sigma{=}0.02$, $r{=}0.2$, $o{=}0.9$, $\theta{=}30^{\circ}$).
  Mean $\pm$ std over 20 trials; best in \textbf{bold}.}
  \label{tab:e1-reference}
  \begin{tabular}{lccc}
    \toprule
    Method & \revt{RE ($^{\circ}$) $\downarrow$} & RMSE ($\times 10^{-3}$) $\downarrow$ & Time (s) \\
    \midrule
    TEASER++~\cite{yang2021teaser}      & \revt{1.66 $\pm$ 1.02}  & \revm{17.1 $\pm$ 7.3} & \textbf{0.09} \\
    FGR~\cite{zhou2016fast}           & \revt{13.00 $\pm$ 13.66} & \rev{126.4 $\pm$ 99.8} & 0.10 \\
    Sparse-ICP~\cite{bouaziz2013sparseicp}    & \revt{27.93 $\pm$ 6.40} & 428.0 $\pm$ 109.0 & \revm{0.12} \\
    CPD~\cite{myronenko2010point}           & \revt{0.33 $\pm$ 0.31} & 3.3 $\pm$ 2.6 & \revm{5.49} \\
    FilterReg~\cite{gao2019filterreg}     & \revt{0.27 $\pm$ 0.15} & 3.6 $\pm$ 1.2 & 0.18 \\
    LSG-CPD~\cite{liu2021lsgcpd}       & \revt{7.63 $\pm$ 8.75} & 77.9 $\pm$ 93.7 & 0.20 \\
    RPOT~\cite{qin2022partialot}          & \revt{0.57 $\pm$ 0.33} & 7.0 $\pm$ 2.6 & \revm{3.96} \\
    \rev{RobOT~\cite{shen2021accurate}} & \rev{5.89 $\pm$ 2.51} & \rev{73.0 $\pm$ 15.1} & \rev{4.97} \\
    \midrule
    \textbf{Sinkhorn-CPD} & \revt{\textbf{0.21 $\pm$ 0.10}} & \textbf{2.6 $\pm$ 0.8} & \revm{4.09} \\
    \bottomrule
  \end{tabular}
\end{table}

\begin{figure*}[pos=t]
  \centering
  \includegraphics[width=\linewidth]{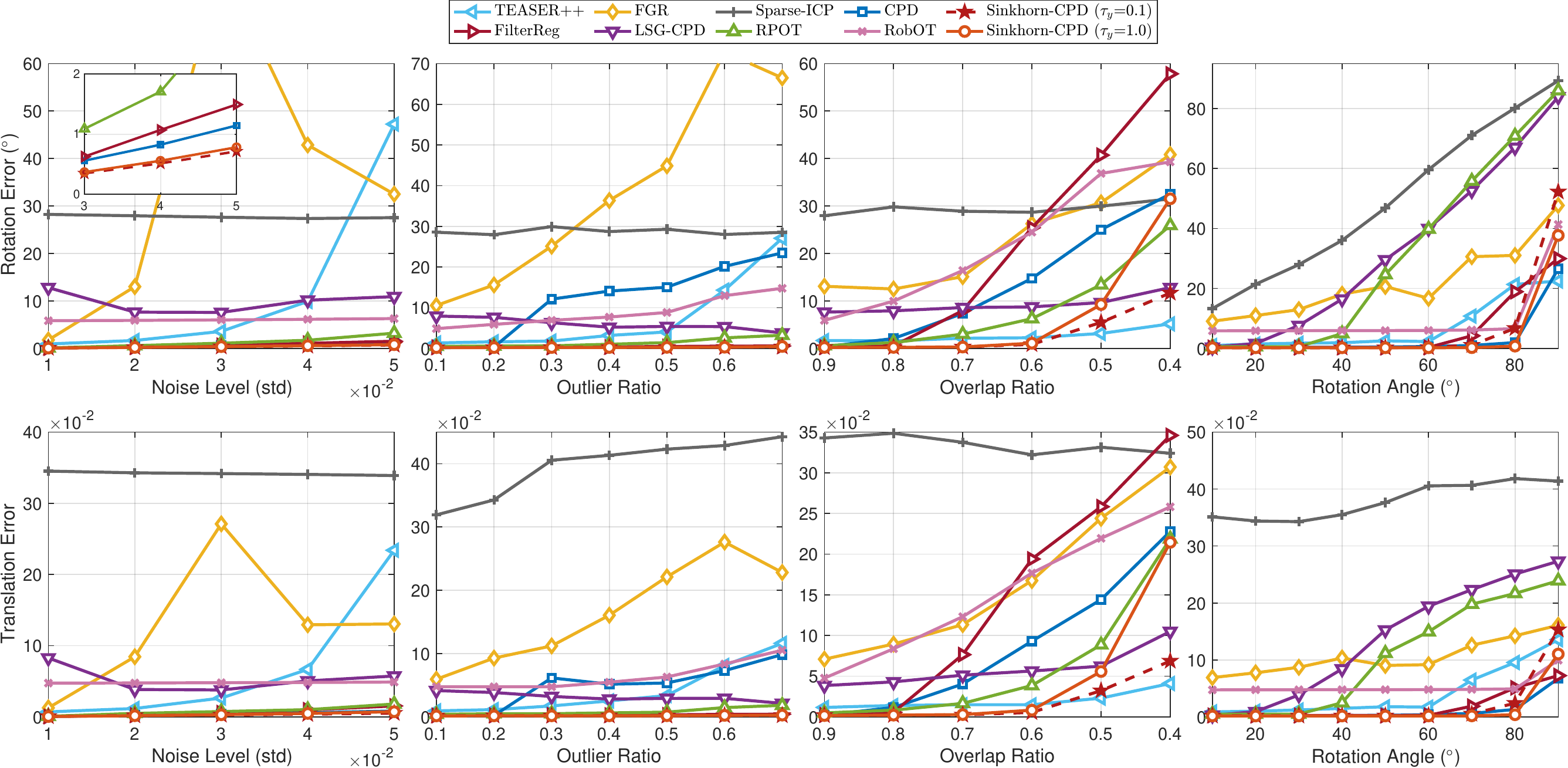} \\
  \makebox[0.25\linewidth]{\footnotesize (a) Noise}%
  \makebox[0.25\linewidth]{\footnotesize (b) Outlier}%
  \makebox[0.25\linewidth]{\footnotesize (c) Overlap}%
  \makebox[0.25\linewidth]{\footnotesize (d) Rotation}
  \caption{Robustness curves on the Stanford Bunny benchmark (mean over 20 trials).
  Top: rotation error (RE); bottom: translation error (TE).
  Sinkhorn-CPD (orange) maintains the lowest or near-lowest error across all conditions.}
  \label{fig:e1-robustness}
\end{figure*}

\noindent\textbf{Robustness curves.}
Fig.~\ref{fig:e1-robustness} reports RE/TE as each factor varies.
Sinkhorn-CPD remains among the best methods across all four axes.
For completeness, per-level RE values are listed in Table~\ref{tab:e1-detailed}.

\noindent\textbf{Noise.}
Sinkhorn-CPD remains stable across all noise levels (RE ${<}\,0.72^{\circ}$ at $\sigma{=}0.05$), while FilterReg degrades beyond $\sigma{=}0.03$. The dual-KL formulation down-weights noisy correspondences through soft marginal constraints.

\noindent\textbf{Outlier.}
CPD fails abruptly beyond $20\%$ outliers (RE: $0.33^{\circ} \to 12.10^{\circ}$). Sinkhorn-CPD keeps RE below $0.41^{\circ}$ up to $70\%$ outliers via symmetric mass rejection.

\noindent\textbf{Partial overlap.}
TEASER++ wins at severe partial overlap ($o{=}0.4$, RE $= 5.16^{\circ}$) via global optimality guarantees. Sinkhorn-CPD achieves the best accuracy for $o \geq 0.6$ (RE ${<}\,0.84^{\circ}$).

\noindent\textbf{Rotation.}
\revt{Sinkhorn-CPD ($\tau_y{=}1.0$) maintains sub-degree errors up to $80^{\circ}$ rotation (RE${=}0.75^{\circ}$), the widest basin among local methods; the more aggressive $\tau_y{=}0.1$ degrades earlier (RE${=}6.70^{\circ}$ at $80^{\circ}$).} This stems from a large initial $\sigma^{2}$ yielding a diffuse Gibbs kernel for coarse capture before automatic annealing.

\subsubsection{Mechanistic Comparison}
\label{sec:e1-mechanism}

\begin{figure}[pos=htp]
  \centering
  \includegraphics[width=\linewidth]{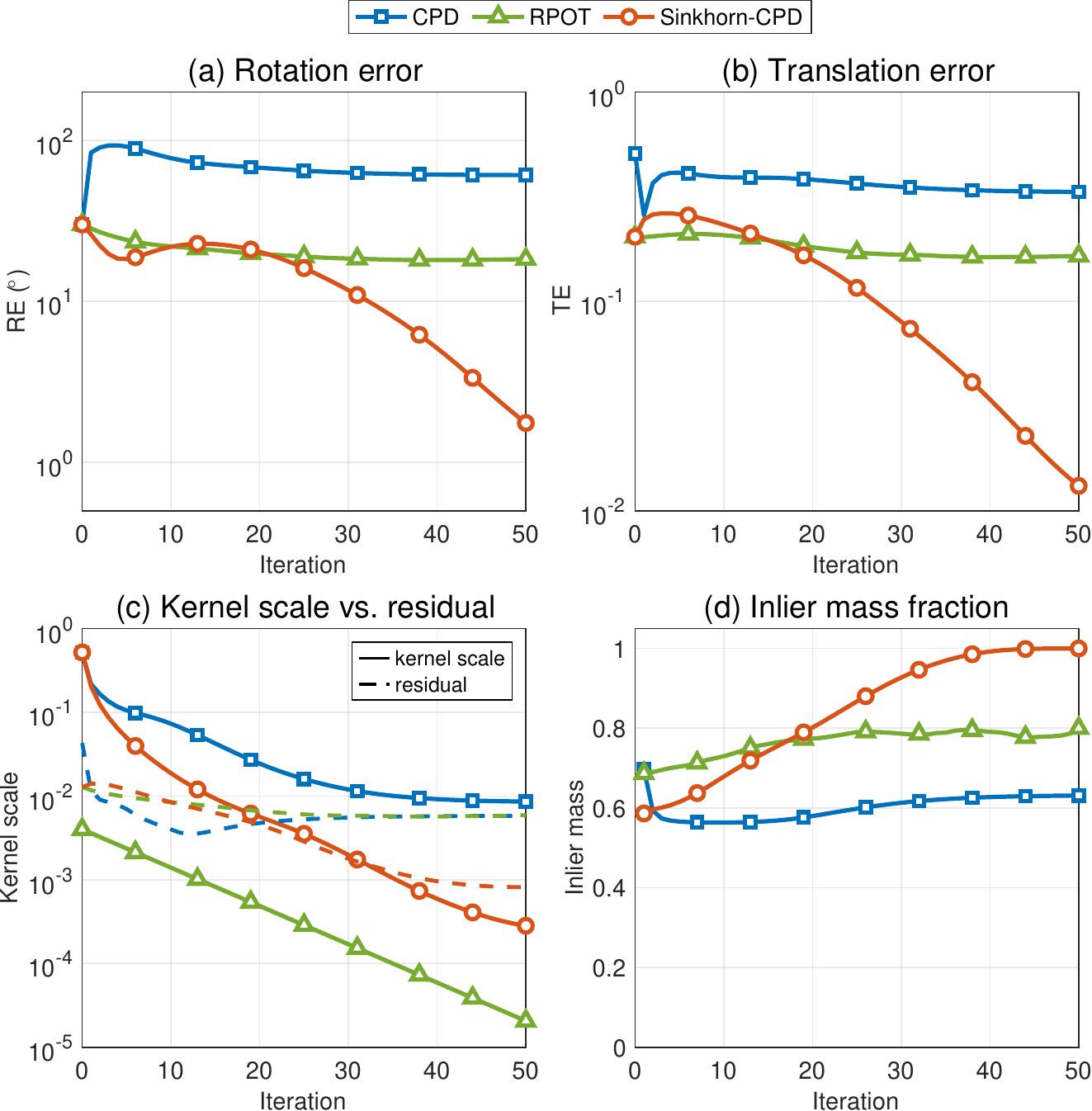}
  \caption{\revt{Per-iteration diagnostics for CPD, RPOT, and Sinkhorn-CPD on a bunny pair ($o{=}0.5$, $20\%$ outliers, $30^{\circ}$ rotation). \textbf{(a,\,b)} Rotation and translation error. \textbf{(c)} Kernel scale (solid) vs.\ mean inlier residual (dashed). \textbf{(d)} Inlier mass fraction.}}
  \label{fig:diag-curves}
\end{figure}

\revt{To understand \emph{why} these differences arise, we instrument CPD, RPOT, and Sinkhorn-CPD on a single bunny pair ($o{=}0.5$, $20\%$ outliers, $30^{\circ}$ rotation) and log per-iteration diagnostics (Fig.~\ref{fig:diag-curves}). The three methods end at $60.9^{\circ}$, $18.3^{\circ}$, and $1.75^{\circ}$ RE respectively, and Fig.~\ref{fig:diag-curves}(c) shows the relationship between kernel scale and alignment residual $\bar{r}^{2}$. CPD's $\sigma^2$ collapses below $\bar{r}^{2}$ within 10 iterations, after which its posteriors lock onto wrong correspondences irreversibly. RPOT decays $\varepsilon$ on a fixed geometric schedule regardless of the actual residual, so it cannot slow down when alignment stalls. Only Sinkhorn-CPD's $\sigma^2$, updated from the transport plan each iteration, tracks $\bar{r}^{2}$ from above: correspondences stay soft until the alignment actually converges. The downstream effect is visible in Fig.~\ref{fig:diag-curves}(d): Sinkhorn-CPD's inlier mass fraction reaches~$1$, while RPOT plateaus at ${\sim}0.80$ and CPD stalls at $0.57$.}

\begin{figure*}[pos=t]
  \centering
  \includegraphics[width=\linewidth]{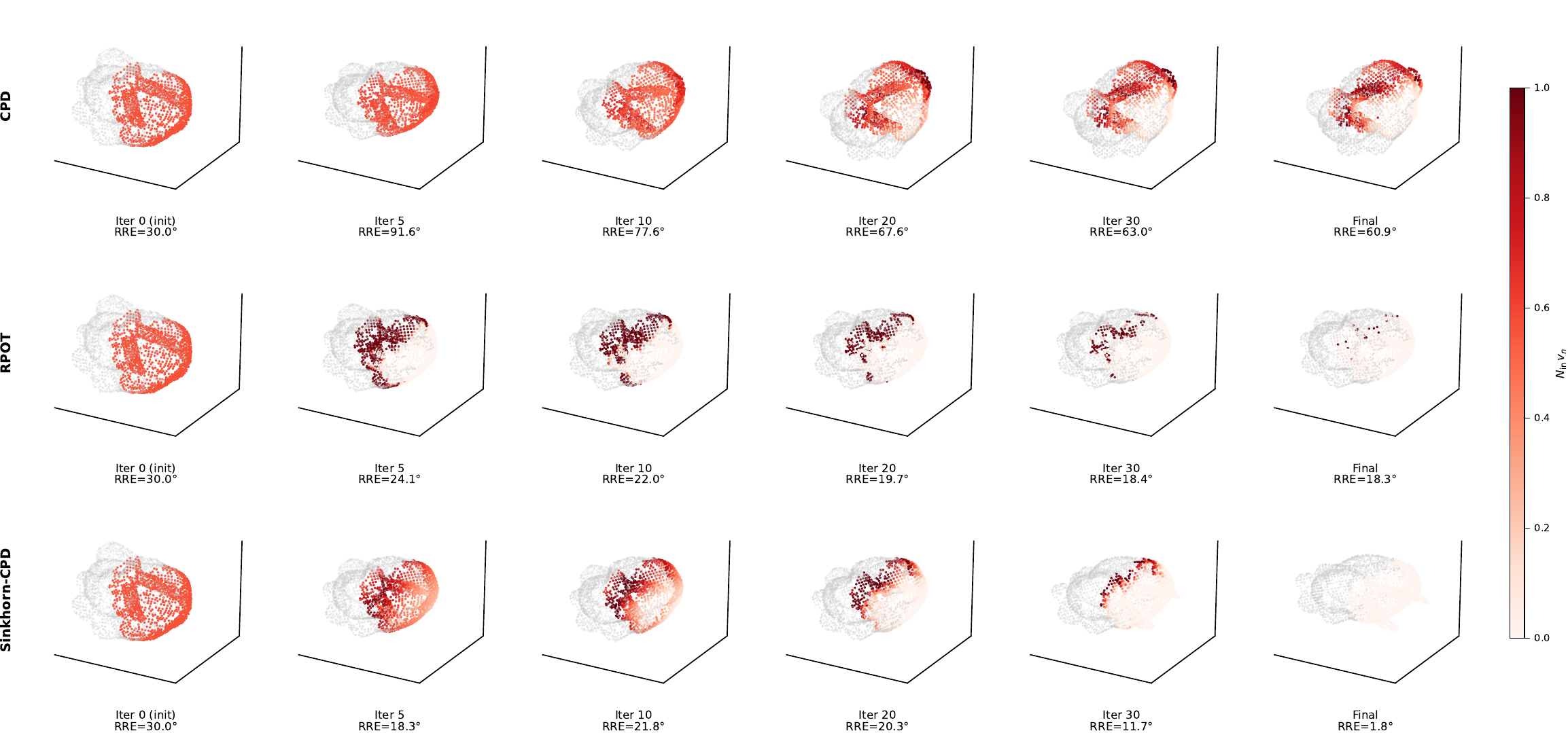}
  \caption{\revt{Per-source-point outlier over-vote $N_{\mathcal{I}}\,v_n$ across iterations for CPD (top), RPOT (middle), and Sinkhorn-CPD (bottom). True inliers are in light gray; outliers are colored by normalized assignment mass ($N_{\mathcal{I}}\,v_n{=}1$ corresponds to one inlier-share).}}
  \label{fig:diag-voteerror}
\end{figure*}

\revt{Fig.~\ref{fig:diag-voteerror} visualizes this spatially: per-source-point outlier over-vote $N_{\mathcal{I}}\,v_n$ across iterations. CPD disperses vote mass over the entire non-overlapping region throughout. RPOT localizes it but never suppresses it, hence the $18.3^{\circ}$ plateau. Sinkhorn-CPD drives outlier mass to near-zero by iteration~30; only the true overlap guides alignment.}

\begin{figure*}[pos=t]
  \centering
  \includegraphics[width=\linewidth]{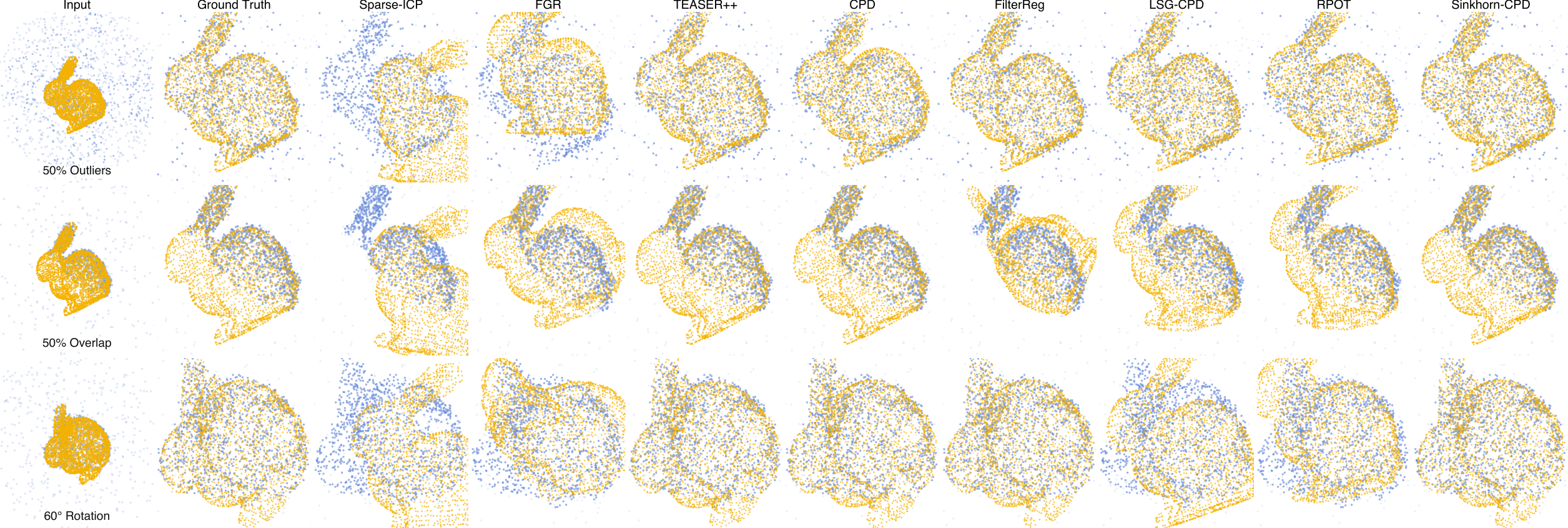}
  \caption{Qualitative comparison on the Stanford Bunny under three challenging scenarios: 50\% outliers (top), 50\% overlap (middle), and $60^{\circ}$ rotation (bottom).
  Sinkhorn-CPD achieves near-zero error across all scenarios.}
  \label{fig:e1-qualitative}
\end{figure*}

\noindent\revt{\textbf{Failure cases and degradation.}}
\revt{Table~\ref{tab:e1-detailed} exposes two failure modes.
At $\theta{=}90^{\circ}$, Sinkhorn-CPD with $\tau_y{=}1.0$ yields $\mathrm{RE}=37.68^{\circ}$ ($52.24^{\circ}$ for $\tau_y{=}0.1$). The diffuse Gibbs kernel has a finite capture range; a $90^{\circ}$ misalignment exceeds it, trapping the algorithm in a local minimum. All local methods share this limitation (CPD: $26.56^{\circ}$, RPOT: $85.98^{\circ}$); only TEASER++, with its certification guarantees, achieves $22.46^{\circ}$.
At $o{=}0.4$, Sinkhorn-CPD with $\tau_y{=}0.1$ achieves $\mathrm{RE}=11.66^{\circ}$, outperformed by TEASER++ ($5.16^{\circ}$). Here the surviving inlier set is too small to reliably condition the Procrustes step, and centroid bias from non-overlapping mass becomes significant.
Overlap degradation is gradual: RE stays below $0.30^{\circ}$ for $o \geq 0.7$, rises to $0.84^{\circ}$ at $o{=}0.6$, jumps to $5.47^{\circ}$ at $o{=}0.5$, and reaches $11.66^{\circ}$ at $o{=}0.4$. At this point non-overlapping mass begins to dominate the transport plan despite dual-KL relaxation.
Fig.~\ref{fig:e1-qualitative} corroborates these degradation trends visually under the three most challenging single-factor settings.}

\begin{figure*}[pos=t]
  \centering
  \includegraphics[width=\linewidth]{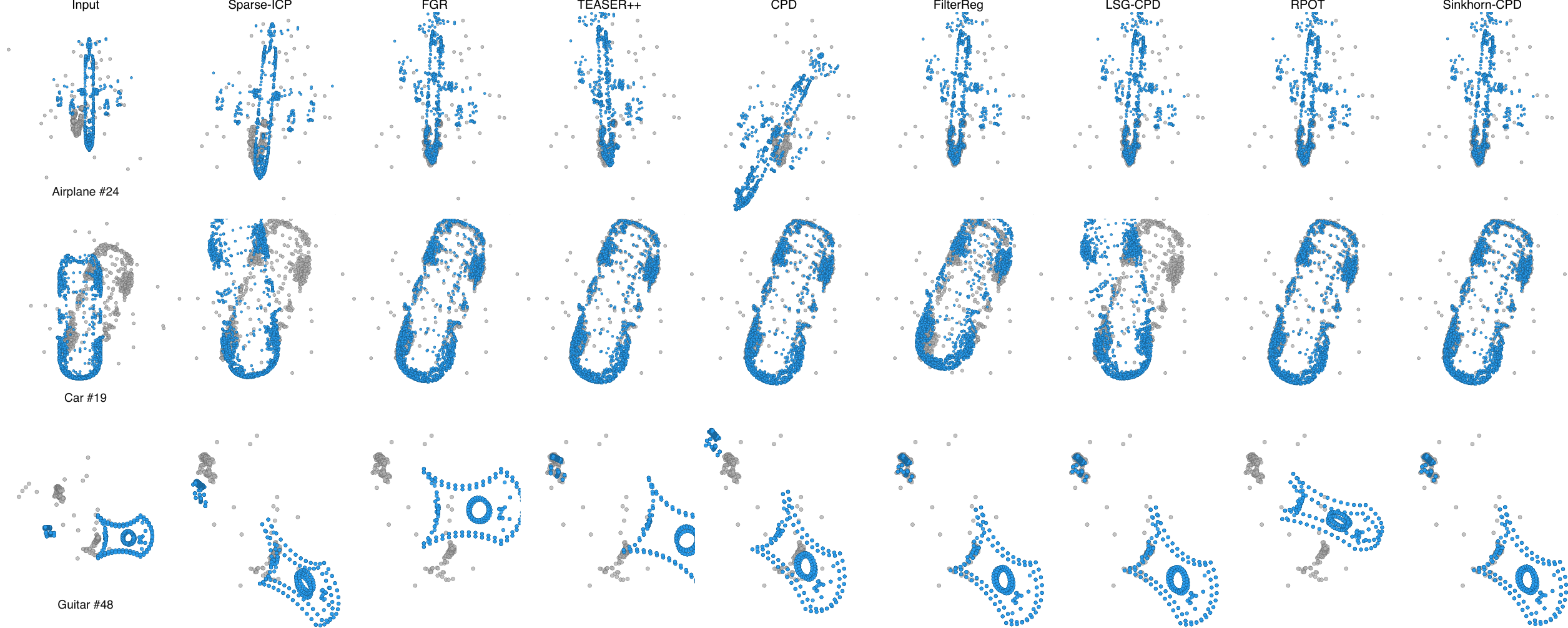}
  \caption{Qualitative registration on ModelNet40 pairs.
  Sinkhorn-CPD achieves consistently low error, while correspondence-based methods (FGR, TEASER++) frequently fail.}
  \label{fig:modelnet-qualitative}
\end{figure*}

\begin{table*}[pos=t]
  \centering
  \footnotesize
  \caption{Mean RE ($^{\circ}$) on the Stanford Bunny across four perturbation axes (20 trials each). Best in \textbf{bold}, second best \underline{underlined}.}
  \label{tab:e1-detailed}
  \resizebox{\textwidth}{!}{%
  \setlength{\tabcolsep}{3.0pt}
  \begin{tabular}{cl cccccccc cc}
    \toprule
    \multirow{2}{*}{Factor} & \multirow{2}{*}{Level}
    & \multirow{2}{*}{TEASER++~\cite{yang2021teaser}}
    & \multirow{2}{*}{FGR~\cite{zhou2016fast}}
    & \multirow{2}{*}{Sparse-ICP~\cite{bouaziz2013sparseicp}}
    & \multirow{2}{*}{CPD~\cite{myronenko2010point}}
    & \multirow{2}{*}{FilterReg~\cite{gao2019filterreg}}
    & \multirow{2}{*}{LSG-CPD~\cite{liu2021lsgcpd}}
    & \multirow{2}{*}{RPOT~\cite{qin2022partialot}}
    & \multirow{2}{*}{\rev{RobOT~\cite{shen2021accurate}}}
    & \multicolumn{2}{c}{\textbf{Sinkhorn-CPD}} \\
    \cmidrule(lr){11-12}
    & & & & & & & & & & $\tau_y{=}0.1$ & $\tau_y{=}1.0$ \\
    \midrule
    \multirow{5}{*}{\textbf{Noise} $\sigma$}
    & 0.01 & \revm{0.92} & \revm{1.76} & 28.22 & 0.09 & \rev{\underline{0.07}} & 12.80 & 0.09 & \rev{5.82} & \underline{0.07} & \textbf{0.06} \\
    & 0.02 & 1.66 & \rev{13.00} & 27.93 & 0.33 & \rev{0.27} & 7.63 & 0.57 & \rev{5.89} & \underline{0.21} & \textbf{0.20} \\
    & 0.03 & \revm{3.57} & \rev{80.86} & 27.62 & 0.56 & \rev{0.63} & 7.57 & 1.09 & \rev{5.99} & \textbf{0.35} & \underline{0.37} \\
    & 0.04 & 9.89 & \rev{42.84} & 27.36 & 0.82 & \rev{1.07} & 10.15 & 1.71 & \rev{6.12} & \textbf{0.51} & \underline{0.56} \\
    & 0.05 & 47.24 & \rev{32.48} & 27.54 & 1.14 & \rev{1.49} & 10.90 & 3.20 & \rev{6.27} & \textbf{0.71} & \underline{0.78} \\
    \midrule
    \multirow{7}{*}{\textbf{Outlier} $r$}
    & 0.1 & \revm{1.33} & \rev{10.54} & 28.54 & \textbf{0.15} & \rev{0.28} & 7.94 & 0.49 & \rev{4.85} & 0.19 & \underline{0.18} \\
    & 0.2 & \revm{1.62} & \rev{15.59} & 27.93 & 0.33 & \rev{0.27} & 7.63 & 0.57 & \rev{5.89} & \underline{0.21} & \textbf{0.20} \\
    & 0.3 & \revm{1.79} & \rev{25.10} & 29.92 & 12.10 & \rev{0.29} & 6.30 & 0.62 & \rev{6.89} & \underline{0.23} & \textbf{0.22} \\
    & 0.4 & \revm{3.18} & \rev{36.34} & 28.71 & 14.08 & \rev{0.34} & 5.20 & 1.02 & \rev{7.70} & \underline{0.26} & \textbf{0.22} \\
    & 0.5 & \revm{4.03} & \rev{44.86} & 29.27 & 15.03 & \rev{0.46} & 5.38 & 1.40 & \rev{8.82} & \underline{0.26} & \textbf{0.25} \\
    & 0.6 & \revm{14.31} & \rev{72.80} & 27.99 & 20.15 & \rev{\underline{0.55}} & 5.34 & 2.58 & \rev{12.95} & \textbf{0.28} & \textbf{0.28} \\
    & 0.7 & \revm{27.06} & \rev{66.48} & 28.52 & 23.46 & \rev{0.58} & 3.80 & 3.22 & \rev{14.76} & \textbf{0.40} & \underline{0.47} \\
    \midrule
    \multirow{6}{*}{\textbf{Overlap} $o$}
    & 0.4 & \textbf{\revm{5.16}} & \rev{40.82} & 31.48 & 32.51 & 57.84 & 12.80 & 25.90 & \rev{39.28} & \underline{11.66} & 31.51 \\
    & 0.5 & \textbf{\revm{3.14}} & \rev{30.69} & 29.96 & 25.00 & 40.70 & 9.64 & 13.39 & \rev{36.84} & \underline{5.47} & 9.23 \\
    & 0.6 & 2.22 & \rev{26.25} & 28.67 & 14.73 & 25.40 & 8.74 & 6.25 & \rev{24.47} & \textbf{0.84} & \underline{1.09} \\
    & 0.7 & 2.14 & \rev{15.10} & 28.89 & \revm{7.35} & \rev{7.95} & 8.59 & 2.99 & \rev{16.40} & \underline{0.30} & \textbf{0.28} \\
    & 0.8 & \revm{1.59} & \rev{12.51} & 29.81 & 2.07 & \rev{0.52} & 7.91 & 1.21 & \rev{9.94} & \underline{0.23} & \textbf{0.20} \\
    & 0.9 & \revm{1.65} & \rev{13.08} & 27.93 & 0.33 & \rev{0.27} & 7.63 & 0.57 & \rev{5.89} & \underline{0.21} & \textbf{0.20} \\
    \midrule
    \multirow{9}{*}{\textbf{Rotation} $\theta$}
    & $10^{\circ}$ & 0.93 & \rev{9.05} & 13.27 & 0.29 & \rev{0.31} & 0.40 & 0.52 & \rev{5.84} & \underline{0.21} & \textbf{0.20} \\
    & $20^{\circ}$ & \revm{1.46} & \rev{10.94} & 21.39 & 0.30 & \rev{0.28} & 1.63 & 0.52 & \rev{5.87} & \underline{0.21} & \textbf{0.20} \\
    & $30^{\circ}$ & \revm{1.72} & \rev{12.93} & 27.93 & 0.33 & \rev{0.26} & 7.63 & 0.57 & \rev{5.89} & \underline{0.21} & \textbf{0.20} \\
    & $40^{\circ}$ & \revm{1.88} & \rev{18.07} & 36.01 & 0.37 & \rev{0.32} & 16.42 & 4.92 & \rev{5.91} & \underline{0.21} & \textbf{0.20} \\
    & $50^{\circ}$ & \revm{2.51} & \rev{20.65} & 46.83 & 0.45 & \rev{0.28} & 29.55 & 24.60 & \rev{5.94} & \underline{0.22} & \textbf{0.20} \\
    & $60^{\circ}$ & \revm{2.26} & \rev{16.68} & 59.60 & 0.61 & \rev{\underline{0.32}} & 39.98 & 39.79 & \rev{5.99} & 0.33 & \textbf{0.20} \\
    & $70^{\circ}$ & \revm{10.77} & \rev{30.58} & 71.04 & \underline{0.96} & \rev{4.19} & 52.62 & 55.79 & \rev{6.12} & 1.06 & \textbf{0.24} \\
    & $80^{\circ}$ & \revm{21.33} & \rev{31.00} & 80.14 & \underline{1.92} & \rev{18.90} & 67.01 & 70.85 & \rev{6.56} & 6.70 & \textbf{0.75} \\
    & $90^{\circ}$ & \textbf{\revm{22.46}} & \rev{47.67} & 89.29 & \underline{26.56} & \rev{29.94} & \revm{83.90} & 85.98 & \rev{41.32} & 52.24 & 37.68 \\
    \bottomrule
  \end{tabular}}
\end{table*}

\subsection{Large-Scale Benchmark: ModelNet40}
\label{sec:e2}

This benchmark tests cross-category generalization without training.
ModelNet40~\cite{wu20153d} contains 40 CAD categories.
We follow the pipeline in \cite{qin2023geotransformer}. 
For each shape we sample $M{=}N{=}2{,}048$ points normalized to the unit sphere.
We generate pairs using random rotations (up to $\pm45^{\circ}$), translations (magnitude $0.5$), Gaussian noise ($\sigma=0.05$), and asymmetric planar cropping (keep ratio $0.9$).
In addition to the \revt{eight} classical baselines, we include RPMNet~\cite{yew2020rpm} and GeoTransformer~\cite{qin2023geotransformer} using official pre-trained weights.

\begin{table*}[pos=t]
  \centering
  \footnotesize
  \caption{ModelNet40 results.
  RR: registration recall ($\mathrm{RE}{<}1^{\circ} \wedge \mathrm{TE}{<}0.1$).
  Best in \textbf{bold}, second best \underline{underlined}.}
  \label{tab:e2-modelnet}
  \begin{tabular}{llccccc}
    \toprule
    Family & Method & RE ($^{\circ}$) $\downarrow$ & TE ($\times 10^{-3}$) $\downarrow$ & RMSE ($\times 10^{-3}$) $\downarrow$ & RR (\%) $\uparrow$ & Time (s) \\
    \midrule
    \multirow{3}{*}{Correspondence}
    & TEASER++~\cite{yang2021teaser}      & \rev{3.70}  & \rev{23.3} & \rev{21.7} & \rev{60.6} & \revm{0.07} \\
    & FGR~\cite{zhou2016fast}           & \rev{5.58}  & \rev{44.7} & \rev{45.7} & \rev{18.0} & \revm{0.08} \\
    & Sparse-ICP~\cite{bouaziz2013sparseicp}    & 18.58 & 253.2 & 268.3 &  1.0 & \revm{0.11} \\
    \midrule
    \multirow{3}{*}{Probabilistic}
    & CPD~\cite{myronenko2010point}           & \rev{1.73}  & \rev{28.6} & \rev{30.0} & \rev{59.9} & \revm{3.03} \\
    & FilterReg~\cite{gao2019filterreg}     & \rev{1.29}  & 16.4 & \rev{16.6} & \underline{\rev{85.5}} & 0.12 \\
    & LSG-CPD~\cite{liu2021lsgcpd}       & 8.14  & 134.8 & 147.2 & \revm{28.8} & \revm{0.15} \\
    \midrule
    \multirow{2}{*}{Learning}
    & RPMNet~\cite{yew2020rpm}        & 1.14  & 16.0 & 16.8 & 59.1 & \underline{0.06} \\
    & GeoTransformer~\cite{qin2023geotransformer} & \underline{0.81} & \textbf{9.0} & \textbf{9.0} & 81.2 & \textbf{0.05} \\
    \midrule
    \multirow{3}{*}{OT-based}
    & RPOT~\cite{qin2022partialot}          & 2.81  & 33.7 & 40.1 & 71.6 & \revm{1.03} \\
    & \rev{RobOT~\cite{shen2021accurate}}   & \rev{5.67} & \rev{87.4} & \rev{93.4} & \rev{5.9} & \rev{0.53} \\
    & \textbf{Sinkhorn-CPD (Ours)} & \textbf{0.64} & \underline{13.5} & \underline{14.3} & \textbf{88.9} & \revm{0.53} \\
    \bottomrule
  \end{tabular}
\end{table*}

As Table~\ref{tab:e2-modelnet} shows, Sinkhorn-CPD achieves the highest registration recall ($88.9\%$) and the lowest mean RE ($0.64^{\circ}$). GeoTransformer achieves a slightly lower TE ($9.0\times 10^{-3}$) but struggles with symmetrically ambiguous shapes, reflected in a lower recall ($81.2\%$). RPMNet achieves only $59.1\%$ recall, highlighting the difficulty learned descriptors face under novel noise and cropping configurations. Sinkhorn-CPD surpasses both deep models on unseen categories by $7.7$ percentage points in recall without any training data.
\revt{Notably, RobOT~\cite{shen2021accurate} achieves only $5.9\%$ recall despite sharing the dual-KL unbalanced OT framework with Sinkhorn-CPD. Its fixed regularization parameter $\varepsilon$ cannot adapt to the diverse noise levels across ModelNet40 categories, whereas Sinkhorn-CPD's data-driven $\sigma^2$ update automatically adjusts the transport sharpness.}
Fig.~\ref{fig:modelnet-qualitative} illustrates qualitative comparisons.

\subsection{Real-World Scan-to-CAD Registration}
\label{sec:e5-real}

We assess practical applicability on a real-world scan-to-CAD dataset of 50 industrial parts that we physically scanned ourselves. The raw scans contain noise, outliers, and partial surface coverage due to occlusion.
We first merged the fragmented scans of each part into a complete point cloud, then registered each scan to its corresponding CAD model via coarse-to-fine alignment (RANSAC followed by ICP refinement, iterated until RMSE convergence). The resulting transformations serve as approximate ground truth.

For this benchmark, both source and target clouds are normalized and downsampled to 4,096 points. 
We generate 1,000 test pairs by applying random initial misalignments to the source scans (rotation up to $30^{\circ}$, translation $\pm 0.1$).

\begin{figure*}[pos=t]
  \centering
  \includegraphics[width=1.0\linewidth]{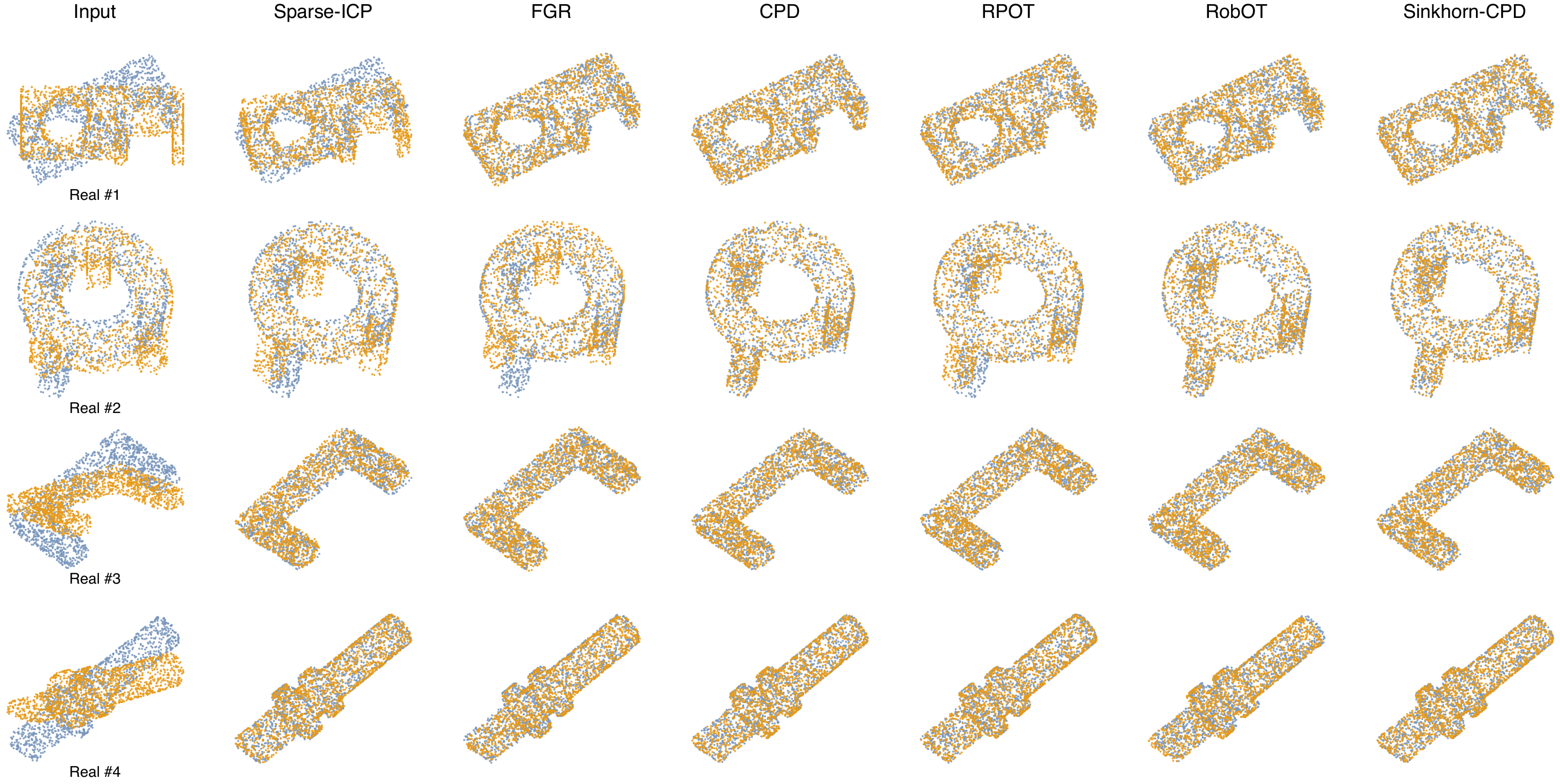}
  \caption{Qualitative scan-to-CAD registration on four representative cases from the real-perturbed CAD dataset.}
  \label{fig:scan2cad-qualitative}
\end{figure*}

Table~\ref{tab:e5-real} shows the results. Feature-based approaches (FGR, TEASER++) fail frequently because smooth, featureless industrial surfaces do not support robust local descriptors. Sinkhorn-CPD achieves the lowest mean RE ($8.52^{\circ}$) and TE ($7.2\times 10^{-3}$), comparable to FilterReg ($8.61^{\circ}$) in RE but with a lower TE and standard deviation.
\revt{RobOT~\cite{shen2021accurate} achieves comparable RE ($8.61^{\circ}$) but substantially higher TE ($28.7\times 10^{-3}$ vs.\ $7.2\times 10^{-3}$), indicating that its fixed $\varepsilon$ yields coarser translational alignment.}
By combining probabilistic closed-form updates with unbalanced optimal transport, Sinkhorn-CPD naturally suppresses CAD-to-scan discrepancies.
Fig.~\ref{fig:scan2cad-qualitative} provides visual results.

\begin{table}[pos=htp]
  \centering
  \footnotesize
  \caption{Real scan-to-CAD benchmark.
  Best in \textbf{bold}.}
  \label{tab:e5-real}
  \begin{tabular}{lccc}
    \toprule
    Method & RE ($^{\circ}$) $\downarrow$ & TE ($\times 10^{-3}$) $\downarrow$ & Time (s) \\
    \midrule
    TEASER++~\cite{yang2021teaser} & 26.94 $\pm$ 54.49 & 48.8 $\pm$ 130.2 & \rev{\textbf{0.15}} \\
    FGR~\cite{zhou2016fast} & 26.40 $\pm$ 49.82 & 40.1 $\pm$ 34.4 & \rev{0.21} \\
    Sparse-ICP~\cite{bouaziz2013sparseicp} & 14.10 $\pm$ 8.27 & 36.7 $\pm$ 25.0 & \rev{0.19} \\
    CPD~\cite{myronenko2010point} & 8.76 $\pm$ 12.38 & 9.1 $\pm$ 14.3 & \rev{10.47} \\
    FilterReg~\cite{gao2019filterreg} & \rev{8.66 $\pm$ 11.73} & 7.4 $\pm$ 11.5 & \rev{0.23} \\
    LSG-CPD~\cite{liu2021lsgcpd} & 11.30 $\pm$ 9.40 & 15.2 $\pm$ 17.2 & \rev{0.49} \\
    RPOT~\cite{qin2022partialot} & 9.28 $\pm$ 10.19 & 8.8 $\pm$ 12.4 & \rev{7.29} \\
    \rev{RobOT~\cite{shen2021accurate}} & \rev{8.61 $\pm$ 12.17} & \rev{28.7 $\pm$ 11.4} & \rev{7.68} \\
    \textbf{Sinkhorn-CPD} & \textbf{8.52 $\pm$ 10.94} & \textbf{7.2 $\pm$ 10.2} & \rev{2.64} \\
    \bottomrule
  \end{tabular}
\end{table}

\subsection{Ablation Studies}
\label{subsec:analysis-ablation}

We validate the key design choices of Sinkhorn-CPD through three ablation studies.

\subsubsection{Sensitivity to KL Penalty Weights}

We perform a grid search over $\tau_x \in \{0.1, 0.5, 1, 2, 5\}$ and $\tau_y \in \{0.1, 0.5, 1, 2, 5\}$ (25 configurations) on the Stanford Bunny reference setting ($\sigma{=}0.02$, $r{=}0.2$, $o{=}0.9$, $\theta{=}30^{\circ}$), running 20 trials per configuration. 
As shown in Fig.~\ref{fig:analysis}(a), the default $\tau_x=\tau_y=1$ resides in a broad, low-error basin, proving the method is relatively insensitive to exact hyperparameter tuning. Setting $\tau \leq 0.1$ degrades accuracy because the marginal constraints become too weak to guide the alignment.

\subsubsection{One-Sided vs.\ Dual KL}

To quantify the benefit of relaxing \emph{both} marginals, we compare four configurations on the outlier experiment ($r \in \{0.1, \ldots, 0.7\}$):
(i)~\emph{Dual-KL} ($\tau_x{=}\tau_y{=}1$): the full Sinkhorn-CPD;
(ii)~\emph{One-sided-Y} ($\tau_x{\to}\infty$, $\tau_y{=}1$): only the source-side marginal is relaxed;
(iii)~\emph{One-sided-X} ($\tau_x{=}1$, $\tau_y{\to}\infty$): only the scan-side marginal is relaxed;
(iv)~\emph{CPD}: the fully one-sided baseline.

Fig.~\ref{fig:analysis}(b) confirms that the dual-KL formulation uniquely survives heavy contamination. One-sided-X handles target outliers well but fails under source outliers, while one-sided-Y mimics the brittle behavior of CPD. Symmetric relaxation is strictly necessary.

\subsubsection{Variance Annealing Mechanism}

We ablate the $\sigma^2$ update mechanism by comparing:
(i)~\emph{Full}: standard Sinkhorn-CPD with automatic $\sigma^2$ updates;
(ii)~\emph{Fixed-$\sigma^2$}: $\sigma^2$ frozen at its initial value (no annealing);
(iii)~\emph{No-$\sigma^2$}: cost matrix uses only squared distances without $\sigma^2$ normalization.

Fig.~\ref{fig:analysis}(c) reports RE across all four perturbation axes.
The full model with automatic annealing consistently outperforms both ablated variants, confirming that variance-driven cost normalization is critical for robust registration.

\revt{Fig.~\ref{fig:diag-curves}(c) provides per-iteration evidence: Sinkhorn-CPD's $\sigma^{2}$ tracks the alignment residual from above, while RPOT's fixed geometric decay ($\varepsilon\leftarrow 0.9\,\varepsilon$) cannot slow down when the residual stays large, and CPD's $\sigma^{2}$ collapses prematurely. Variance-driven annealing is what makes the difference.}

\subsubsection{\revt{Effect of $\tau_y$ on Overlap Robustness}}

\begin{figure*}[pos=t]
  \centering
  \includegraphics[width=\linewidth]{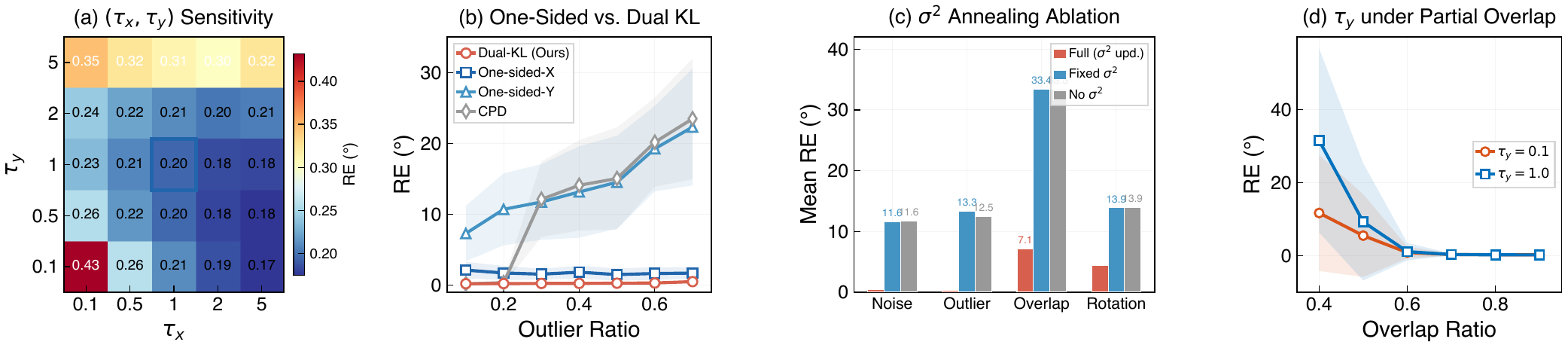}
  \caption{Ablation studies on the Bunny benchmark.
  (a) Mean RE over a grid of $(\tau_x,\tau_y)$.
  (b) Outlier robustness for different marginal relaxations.
  (c) Effect of variance-driven annealing.
  \revt{(d) $\tau_y{=}0.1$ vs.\ $\tau_y{=}1.0$ under varying overlap: the stronger target-side penalty trades convergence basin width for partial-overlap robustness.}}
  \label{fig:analysis}
\end{figure*}

\revt{Fig.~\ref{fig:analysis}(d) compares the two $\tau_y$ settings under varying overlap. At high overlap ($o \geq 0.7$), both achieve sub-degree RE; as overlap decreases, $\tau_y{=}0.1$ degrades more gracefully ($\mathrm{RE}=11.7^{\circ}$ at $o{=}0.4$) than $\tau_y{=}1.0$ ($\mathrm{RE}=31.5^{\circ}$), because the stronger target-side penalty suppresses non-overlapping mass more effectively. In practice, this is the trade-off when choosing $\tau_y$: partial-overlap robustness versus convergence basin width (Section~\ref{sec:e1}).}

\subsection{Computational Cost}
\label{subsec:runtime}

Tables~\ref{tab:e1-reference} and~\ref{tab:e2-modelnet} detail the wall-clock execution times. On the Bunny benchmark ($N=3{,}000$), Sinkhorn-CPD runs in \revt{4.09}\,s, matching the efficiency of RPOT (\revt{3.96}\,s) and running faster than CPD (\revt{5.49}\,s). On ModelNet40 ($N=2{,}048$), it requires \revt{0.53}\,s.
The computational bottleneck is the $\mathcal{O}(MN)$ pairwise distance calculation. Because Sinkhorn-CPD inherits CPD's data-driven annealing, it typically converges in 15--30 outer iterations without overshooting. The $L=20$ inner Sinkhorn iterations introduce minimal overhead relative to the distance computation and closed-form Procrustes updates.

\section{Assumptions and Limitations}
\label{sec:limitations}

\revt{Sinkhorn-CPD inherits several modeling assumptions from CPD that bound its applicability:}

\revt{\textbf{Isotropic, homogeneous noise.}
A single scalar $\sigma^2$ models noise uniformly across all directions and points. Extending to per-point or directional covariances would require replacing the scalar Sinkhorn scaling with a matrix form.}

\revt{\textbf{Known scale.}
Only rotation and translation are estimated; source and target must share the same metric scale.}

\revt{\textbf{Gaussian cost.}
The squared-distance cost assumes Gaussian inlier noise. Dual-KL relaxation provides empirical robustness up to $70\%$ outliers (Section~\ref{sec:e1}), but for heavy-tailed inlier noise a robust cost would be more appropriate.}

\revt{\textbf{Local convergence.}
Sinkhorn-CPD is a local method: convergence is reliable up to $80^{\circ}$ initial rotation ($\mathrm{RE}{=}0.75^{\circ}$) but fails at $90^{\circ}$ ($\mathrm{RE}{=}37.68^{\circ}$). Under poor initialization, the transport plan can concentrate on a wrong subset, triggering premature $\sigma^2$ collapse. A global initializer (RANSAC, TEASER++) resolves this in practice.}

\revt{\textbf{Partial overlap.}
Under severe partial overlap ($o \leq 0.5$), non-overlapping mass biases the weighted centroid despite dual-KL relaxation. Degradation is gradual: $\mathrm{RE} < 0.84^{\circ}$ for $o \geq 0.6$, rising to $11.66^{\circ}$ at $o{=}0.4$.}

\section{Conclusion}
\label{sec:conclusion}

We proposed \emph{Sinkhorn-CPD}, which formulates rigid registration as an unbalanced entropic OT problem with dual KL penalties on both marginals while retaining CPD's Gaussian log-likelihood cost.
The algorithm alternates generalized Sinkhorn scaling with closed-form Procrustes and variance updates, achieving symmetric outlier rejection without manual temperature schedules or training data.

On the Stanford Bunny benchmark, Sinkhorn-CPD with $\tau_y{=}0.1$ maintains $\mathrm{RE} < 0.41^{\circ}$ at $70\%$ outlier ratio, where CPD collapses beyond $20\%$.
On ModelNet40 with unseen categories, it achieves $88.9\%$ registration recall, surpassing GeoTransformer ($81.2\%$) without any training.

Sinkhorn-CPD inherits the $\mathcal{O}(MN)$ complexity and local convergence basin of CPD, which limits its scalability to very large point sets and its capture range for extreme initial misalignments.
We plan to explore scalable kernel approximations (\eg, Nystr\"om) to reduce the pairwise distance bottleneck, and to extend the framework to non-rigid registration by incorporating deformation models.

\section*{Acknowledgments}

This study was supported by the Beijing Natural Science Foundation (No. Z240002) and the National Key Research and Development Program of China (No. 2024YFA1013102).

\small
\bibliography{mybibfile}

\appendix

\section{Proof of Theorem \ref{thm:equivalence}}
\label{app:cpd-ot}

We show that optimizing the CPD E-step is equivalent to solving a one-sided entropic OT problem. 
Assuming uniform mixture weights and no explicit uniform outlier component ($w=0$) to isolate the geometric core, the variational free energy expands as:
\begin{equation}
  \mathcal{F}_{\mathrm{VFE}}(\mathbf{P}) = \sum_{m,n} P_{mn} \tilde{C}_{mn} + \sum_{m,n} P_{mn} \log P_{mn} + C',
\end{equation}
where $C'$ is a constant, and the responsibility matrix $\mathbf{P} \in \mathbb{R}_+^{M \times N}$ satisfies the strict column constraint $\mathbf{P}^{\top}\mathbf{1}_M = \mathbf{1}_N$. This implies the total mass is $\sum_{m,n} P_{mn} = N$.

Substituting the Gaussian negative log-likelihood $\tilde{C}_{mn} = \frac{D_{mn}}{2\sigma^2} + \frac{D}{2}\log(2\pi\sigma^2)$ where $D_{mn} = \|\mathbf{x}_n - T_{\boldsymbol{\theta}}(\mathbf{y}_m)\|^2$, we obtain:
\begin{align}
  \mathcal{F}_{\mathrm{VFE}} &= \frac{1}{2\sigma^2} \sum_{m,n} P_{mn} D_{mn} + \frac{D}{2}\log(2\pi\sigma^2) \sum_{m,n} P_{mn} \nonumber \\
  &\quad + \sum_{m,n} P_{mn} \log P_{mn} + C'.
\end{align}
Because the total mass is fixed to $N$, the second term acts as a constant with respect to $\mathbf{P}$. Factoring out $\frac{1}{2\sigma^2}$, minimizing $\mathcal{F}_{\mathrm{VFE}}$ over $\mathbf{P}$ is identical to solving:
\begin{equation}
  \min_{\substack{\mathbf{P} \geq \mathbf{0} \\ \mathbf{P}^{\top}\mathbf{1}_M = \mathbf{1}_N}} \langle \mathbf{D}, \mathbf{P} \rangle + 2\sigma^2 \sum_{m,n} P_{mn} \log P_{mn}.
\end{equation}
Since the total mass is fixed to $N$, we have $\sum P_{mn}\log P_{mn} = \mathcal{H}(\mathbf{P}) + N$, so replacing $\sum P_{mn}\log P_{mn}$ with $\mathcal{H}(\mathbf{P})$ shifts the objective by the constant $2\sigma^2 N$ and preserves the minimizer. This matches the formulation of entropic optimal transport with cost $\mathbf{D}$, regularization strength $\varepsilon = 2\sigma^2$, a fixed target marginal, and a completely free source marginal. \hfill$\square$

\section{Derivation of the Variance Update}
\label{app:variance-derivation}

We compute the optimal variance $\sigma^2$ by minimizing the data-fit term in the objective $\mathcal{J}$ during the M-step. 
Let $E = \sum_{m,n} \Gamma_{mn} \|\mathbf{x}_n - (\mathbf{R}\mathbf{y}_m+\mathbf{t})\|^2$ be the total weighted squared residual. The cost term expands to:
\begin{equation}
  \langle \mathbf{C}, \mathbf{\Gamma} \rangle = \frac{E}{2\sigma^2} + \frac{D\gamma}{2}\log(2\pi\sigma^2),
\end{equation}
where $\gamma = \sum_{m,n} \Gamma_{mn}$ is the total transported mass.
Taking the partial derivative with respect to $\sigma^2$ yields:
\begin{equation}
  \frac{\partial}{\partial \sigma^2} \langle \mathbf{C}, \mathbf{\Gamma} \rangle = -\frac{E}{2\sigma^4} + \frac{D\gamma}{2\sigma^2}.
\end{equation}
Setting the derivative to zero and multiplying by $2\sigma^4 / (D\gamma)$ isolates $\sigma^2$:
\begin{equation}
  \sigma^2 = \frac{E}{D\gamma}.
\end{equation}
This establishes the closed-form variance update shown in Eq.~\eqref{eq:sinkhorn-cpd-sigma}. \hfill$\square$

\section{Baseline Hyperparameters}
\label{app:baseline-params}

\begin{table}[pos=htp]
  \centering
  \footnotesize
  \caption{Baseline hyperparameters. All parameters follow the defaults in the respective implementations.}
  \label{tab:baseline-params}
  \resizebox{\columnwidth}{!}{%
  \setlength{\tabcolsep}{4pt}
  \begin{tabular}{llc}
    \toprule
    Method & Parameters & Lang.\ \\
    \midrule
    TEASER++~\cite{yang2021teaser}
      & FPFH, voxel\,$=$\,0.05, noise bnd\,$=$\,0.05
      & C++ \\
    FGR~\cite{zhou2016fast}
      & FPFH, voxel\,$=$\,0.05, corr.\ dist\,$=$\,0.075
      & C++ \\
    Sparse-ICP~\cite{bouaziz2013sparseicp}
      & \revt{$p=0.5$}
      & C++ \\
    CPD~\cite{myronenko2010point}
      & \revt{$w=0.5$}
      & Py/CuPy \\
    FilterReg~\cite{gao2019filterreg}
      & $w=0.1$
      & Python \\
    LSG-CPD~\cite{liu2021lsgcpd}
      & $w=0.5$
      & MATLAB \\
    RPOT~\cite{qin2022partialot}
      & $\varepsilon_0=0.004$, ann.\,$\times$\,0.9, $\beta_\mathrm{m}=0.8$
      & MATLAB \\
    \rev{RobOT~\cite{shen2021accurate}}
      & \revt{blur\,$=$\,0.1, reach\,$=$\,1.0, scaling\,$=$\,0.9}
      & \rev{PyTorch} \\
    RPMNet~\cite{yew2020rpm}
      & official pre-trained weights
      & PyTorch \\
    GeoTransformer~\cite{qin2023geotransformer}
      & official pre-trained weights
      & PyTorch \\
    \bottomrule
  \end{tabular}}
\end{table}

Table~\ref{tab:baseline-params} lists the key hyperparameters for each baseline. Unless noted otherwise, all values follow the defaults in the respective official implementations. \revt{All iterative methods share a maximum of 50 outer iterations.}
For TEASER++ and FGR, input correspondences are established via FPFH descriptors~\cite{rusu2009fast} computed with Open3D, using a search radius of $6\times$ (TEASER++) or $5\times$ (FGR) the voxel size. TEASER++ additionally applies mutual nearest-neighbor filtering.
\revt{For RobOT we omit the FPFH front-end used in the official demo and run optimal transport directly on raw 3D coordinates, matching all other coordinate-only baselines.}
\revt{The Sparse-ICP $p=0.5$ value is compiled into the C++ binary shipped and is not exposed at the Python wrapper level.}
RPMNet and GeoTransformer use their official pre-trained weights without fine-tuning. Both are evaluated on the same data pairs as the classical methods with identical random seeds.

\section{\revt{Numerical Stabilization}}
\label{app:numerical}

\revt{Two safeguards are applied in the Sinkhorn-CPD implementation. (i)~A floor $\sigma^2 \ge 10^{-8}$ is enforced after each variance update. Under low-noise, low-outlier configurations the inlier residuals can shrink within a few iterations, and the unconstrained closed-form $\sigma^2$ would otherwise approach $0$ and trigger numerical overflow in the Gibbs kernel $\exp(-C/(2\sigma^2))$. (ii)~Inside the generalized Sinkhorn iterations, divisions $\mathbf{b}\oslash(\mathbf{K}\mathbf{v})$ and $\mathbf{a}\oslash(\mathbf{K}^{\top}\mathbf{u})$ use a small additive $\epsilon=10^{-12}$ in the denominator to avoid zero-mass edge cases. Both constants are well below the precision of all reported metrics and do not affect convergence behavior.}

\section{Code and Data Availability}
\label{app:code}

The complete source code is available at \url{https://github.com/Theigrams/SinkhornCPD}. The repository contains:
\begin{itemize}[leftmargin=1.5em, topsep=2pt, itemsep=1pt]
  \item Core algorithm: Sinkhorn-CPD.
  \item Baseline algorithms: CPD, RPOT, FGR, FilterReg, TEASER++, Sparse-ICP, LSG-CPD, RobOT.
  \item Experiment scripts: reproducing all configurations in Tables~\ref{tab:e1-reference}--\ref{tab:e2-modelnet} with fixed random seeds.
  \item Data: Stanford Bunny (${\sim}$30\,MB); ModelNet40 test pairs (227\,MB).
\end{itemize}

\section{\revt{Comparison with OT-Based Methods}}
\label{app:ot-comparison}

\revt{\textbf{RobOT}~\cite{shen2021accurate} (fixed $\varepsilon$, fixed $\rho$, dual-KL):
\begin{equation*}
  \min_{\mathbf{\Gamma} \geq \mathbf{0}}\; \bigl\langle \tfrac{1}{2}\|\mathbf{x}_n{-}T(\mathbf{y}_m)\|^2,\, \mathbf{\Gamma} \bigr\rangle
  + \varepsilon\, \mathcal{H}(\mathbf{\Gamma})
  + \rho\,\KL(\mathbf{\Gamma}\mathbf{1}\|\mathbf{b})
  + \rho\,\KL(\mathbf{\Gamma}^{\!\top}\mathbf{1}\|\mathbf{a}).
\end{equation*}}

\revt{\textbf{RPOT}~\cite{qin2022partialot} (decaying $\varepsilon$, hard partial constraints):
\begin{equation*}
  \min_{\mathbf{\Gamma} \geq \mathbf{0}}\; \langle \|\mathbf{x}_n{-}T(\mathbf{y}_m)\|^2,\, \mathbf{\Gamma} \rangle
  + \varepsilon\, \mathcal{H}(\mathbf{\Gamma}),
  \quad \mathbf{\Gamma}\mathbf{1} \leq \mathbf{b},\;
  \mathbf{\Gamma}^{\!\top}\mathbf{1} \leq \mathbf{a},\;
  \langle \mathbf{\Gamma}, \mathbf{1}\rangle = \beta_m.
\end{equation*}}

\revt{\textbf{Sinkhorn-CPD} (adaptive $\sigma^2$, dual-KL):
\begin{equation*}
  \min_{\mathbf{\Gamma} \geq \mathbf{0}}\; \bigl\langle \tfrac{\|\mathbf{x}_n{-}T(\mathbf{y}_m)\|^2}{2\sigma^2} + \tfrac{D}{2}\!\log(2\pi\sigma^2),\, \mathbf{\Gamma} \bigr\rangle
  + \mathcal{H}(\mathbf{\Gamma})
  + \tau_x\,\KL(\mathbf{\Gamma}^{\!\top}\mathbf{1}\|\mathbf{a})
  + \tau_y\,\KL(\mathbf{\Gamma}\mathbf{1}\|\mathbf{b}).
\end{equation*}}

\revt{The central issue is \emph{cost-scale drift}. RobOT and RPOT both use the raw squared distance $\|\mathbf{x}{-}T(\mathbf{y})\|^2$, whose magnitude drops by orders of magnitude as registration progresses. All regularization coefficients---entropy weight $\varepsilon$, KL penalty $\rho$, or partial-mass threshold $\beta_m$---are calibrated against the initial cost scale. As the cost shrinks, these coefficients become progressively mismatched: in RobOT, the fixed $\varepsilon$ and $\rho$ over-regularize near convergence; in RPOT, the geometric decay $\varepsilon \leftarrow 0.9\,\varepsilon$ is blind to the actual residual, forcing premature sharpening when alignment stalls and unnecessarily slow convergence when alignment is easy. Sinkhorn-CPD resolves this by normalizing the cost with the adaptive $\sigma^2$: as residuals shrink, $\sigma^2$ shrinks proportionally, so the ratios between cost, entropy, and KL terms remain stable throughout optimization. This is why a single $\tau{=}1$ works across all experimental settings without retuning.}

\end{document}